\documentclass{article}

\usepackage[final]{corl_2026}    

\usepackage{amsmath, amssymb}
\usepackage{mathtools}
\usepackage{float}
\usepackage{graphicx}
\usepackage{wrapfig}
\usepackage{hyperref}

\title{Optimal Control with Learned Critics under Unmodeled State Dependencies}

\author{Philipp Schoch, Markus Ryll \\
Autonomous Aerial Systems Lab\\
Technical University of Munich\\
\texttt{\{philipp.schoch,markus.ryll\}@tum.de} \\
}

\begin{document}
\maketitle

\begin{abstract}
Model Predictive Control (MPC) provides a structured and constraint-aware mechanism for decision-making, but its reliance on optimization-friendly analytical dynamics models limits its use in tasks with contacts and other hard-to-model state dependencies. Model-free reinforcement learning avoids explicit modeling assumptions but typically requires large amounts of interaction data. We present a learning-based MPC framework that combines the data efficiency and structure of local model-based planning with learned components that compensate for incomplete dynamics and finite-horizon myopia. The method augments a nominal analytical model with a residual dynamics network that learns missing state-dependent effects from data and combines the resulting planner with a learned action-value critic that injects long-horizon MDP structure into the local iLQR optimization. To make this practical at reinforcement-learning scale, we develop a GPU-accelerated batched iLQR solver that evaluates learned dynamics and critic networks inside the optimal-control loop and solves thousands of trajectory-optimization problems in parallel. The complete system is integrated into a robotics simulator, enabling scalable model-based reinforcement learning under incomplete dynamics. Experiments on biased and incompletely modeled control tasks show that the approach improves closed-loop control performance while preserving the model-based structure needed for efficient constrained trajectory optimization.
\end{abstract}


\section{Introduction}
\label{sec:intro}

Computing optimal control policies for agents acting in stochastic environments is a central challenge in sequential decision making.
Formally, this amounts to solving a Markov Decision Process (MDP)~\cite{SuttonBarto}: finding a policy that maximizes expected cumulative reward over potentially infinite horizons.
The field has produced a rich landscape of approaches, broadly organized into trajectory optimization, model-free reinforcement learning (RL), and model-based reinforcement learning methods~\cite{reiter25survey}.

Model Predictive Control (MPC) performs trajectory optimization in a receding-horizon fashion, enabling precise control and explicit satisfaction of state and input constraints. However, MPC requires an optimization-friendly analytical dynamics model, often difficult or impossible to derive for systems involving contacts or other highly nonlinear effects, and breaks down if relevant state variables remain unmodeled. Contact-rich tasks are a representative case, because making and breaking contact introduces discontinuous phase boundaries over which gradient information can become uninformative~\cite{FOGradients}, while classical multi-phase or mixed-integer formulations lead to large nonlinear programs that are brittle and rarely compatible with real-time replanning~\cite{GCS_shortest_path}.
A second limitation is the finite horizon itself. MPC applies a finite-horizon surrogate to an infinite-horizon MDP, so a horizon that is too short can make the controller greedy, trap it in local minima, or prevent task completion when intermediate subgoals must be reached before the final objective becomes attainable~\cite{Lowrey2018}.
A third limitation is that standard MPC formulations operate on explicit state variables and analytical models, making it nontrivial to use high-dimensional sensor observations such as rgb or depth images directly inside the planner.

Model-free RL sidesteps these limitations: learned policies are cheap at inference and can consume high-dimensional observations.
However, model-free methods are notoriously sample-inefficient, scale poorly with state-space dimensionality, and struggle with long-horizon tasks where reward-sparsity impedes exploration~\cite{SuttonBarto,Aubret2019ASO}.

Model-based RL promises to combine the data efficiency of planning with the flexibility of learning~\cite{reiter25survey}.
A compelling paradigm in this space is to use MPC as a critic: the planner provides a structured, locally accurate estimate of the action-value function, which can be used to bootstrap a global $Q$-approximator which after training supplies long-horizon guidance back to the MPC, allowing even to correct for mismatches in the dynamics model. \cite{Bhardwaj2020MPQ} demonstrates this in the setting of gradient-free sampling-based control.
A key remaining challenge is that MPC critics still require an accurate dynamics model, a requirement that fails whenever relevant state variables lack a tractable analytical description, e.g.\ object contact dynamics in manipulation.

In this work, we develop a learning-augmented constrained MPC framework that embeds two learned components directly into the planner. A residual dynamics model augments the nominal analytical model and exposes otherwise unmodeled state dependencies to optimization, while a learned critic injects global MDP information into the finite-horizon objective to reduce myopia and improve guidance under multi-phase cost-structure; unlike in sampling-based MPC~\cite{Bhardwaj2020MPQ}, the critic enters the Riccati recursion through its first- and second-order derivatives. Following the DDP formulation, we realize this approach with a GPU-accelerated batched ALM-iLQR solver that integrates both learned components inside the optimal-control loop and scales to thousands of parallel environments, enabling efficient model-based reinforcement learning.

\noindent\textbf{Core contributions:} (i)~a constrained MPC formulation that combines residual dynamics learning with a learned critic, and (ii)~a learning-based DDP-style solver that makes this formulation practical for model-based reinforcement-learning, scaling to thousands of parallel environments.

\section{Preliminaries}
\label{sec:prelim}

We consider a discrete-time MDP $\mathcal{M} = (\mathcal{S}, \mathcal{A}, l, P, \gamma, p_0)$~\cite{SuttonBarto}, where $\mathcal{S}$ is the state space, $\mathcal{A}$ the action space, $l(x,u)$ a stage cost, $P$ a stochastic transition model with $x_{k+1} \sim P(\cdot \mid x_k, u_k)$, $\gamma \in (0,1]$ a discount factor, and $p_0$ the initial state distribution.
Let $\mathbb{P}^{\pi}_{\mathcal{M}}(\cdot \mid x,u)$ denote the trajectory distribution induced by $(x_0,u_0)=(x,u)$ and following $\pi$ thereafter.
The action-value and value functions are $Q^\pi_\mathcal{M}(x,u) = \mathbb{E}_{\mathbb{P}^{\pi}_{\mathcal{M}}(\cdot \mid x,u)}\!\left[\sum_{i=0}^{\infty} \gamma^i l(x_i, u_i)\right]$ and $V^\pi_\mathcal{M}(x) = \mathbb{E}_{u \sim \pi(\cdot \mid x)}\!\left[Q^\pi_\mathcal{M}(x,u)\right]$.
The control objective is $\pi^* = \arg\min_{\pi}\, \mathbb{E}_{x \sim p_0}\!\left[V^\pi_\mathcal{M}(x)\right]$.
For any horizon $H \geq 2$, the action-value function admits the finite-horizon decomposition
\begin{equation}
    \label{eq:MPC_objective}
    Q^\pi_\mathcal{M}(x,u)
    =
    \mathbb{E}_{\mathbb{P}^{\pi}_{\mathcal{M}}(\cdot \mid x,u)}\!\left[
    \sum_{i=0}^{H-2} \gamma^i l(x_i, u_i)
    + \gamma^{H-1} Q^\pi_\mathcal{M}(x_{H-1}, u_{H-1})
    \right].
\end{equation}
Classical MPC closes this recursion with a terminal cost, $Q^\pi_\mathcal{M}(x_{H-1}, u_{H-1}) \approx l_T(x_{H-1}, u_{H-1})$~\cite{MAYNE_DDP}.
While sufficient when the terminal cost is well designed, this local approximation may miss the global structure of the MDP, especially when the objective requires reaching an intermediate subgoal~\cite{Lowrey2018}.
In the following we work with deterministic planned trajectories to simplify notation; the stochastic setting is retained by taking expectations over the induced trajectory distribution.

The paper is structured as follows: Section~\ref{sec:method} introduces the blended-critic objective and explains how learning of the residual dynamics function and the learned critic integrates into the framework. Section~\ref{sec:solver} describes the iLQR solver and the specific realization of the blended-critic objective. Section~\ref{sec:experiments} evaluates the method on a set of simulated control tasks, and Sections~\ref{sec:limitations} and~\ref{sec:conclusion} discuss the method's shortcomings and give an outlook on potential extensions.
\section{Optimal Control Objective}
\label{sec:method}

\subsection{Blending the Learned Critic}
\label{sec:blendedCritic_objective}

As established in Eq.~\eqref{eq:MPC_objective}, MPC estimates $Q^\pi_\mathcal{M}$ by rolling out a local trajectory.
However, the accuracy of this estimate depends heavily on the length of the rollout and well-designed terminal cost function.
A global learned critic $\hat{Q}_\theta(x,u)$, parameterized by a neural network with weights $\theta$, can in principle capture the full MDP structure, but suffers from the usual approximation errors of neural networks, making it both hard to train and hard to integrate into a planner (cf. Section~\ref{par:value_ensembles}, cf.~\cite{SuttonBarto}).
Following~\cite{Bhardwaj2020MPQ}, we therefore blend both estimates along the $H$-step planned trajectory $(x_i, u_i)_{i=0}^{H-1}$. As can be seen later, this will allow us to trade off bias in the nominal model with approximation bias in the learned critic, while reducing variance in the value targets during training. 

Following the general Bellman recursion $Q_i = l_i + \gamma Q_{i+1}$, for any $\lambda \in [0,1]$, we define the blended Q-estimate recursively for $i = 0, \dots, H-2$ as
\begin{equation}
    \label{eq:blended_recursion}
    Q^{\lambda}(x_i, u_i)
    =
    (1-\lambda)\hat{Q}(x_i, u_i)
    + \lambda\bigl(l(x_i, u_i) + \gamma\, Q^{\lambda}(x_{i+1}, u_{i+1})\bigr),
\end{equation}
but contrary to~\cite{Bhardwaj2020MPQ}, we terminate the recursion at the last step by $Q^{\lambda}(x_{H-1}, u_{H-1}) = (1-\lambda)\hat{Q}(x_{H-1}, u_{H-1}) + \lambda\, l(x_{H-1}, u_{H-1})$, as a weighted average between the finite horizon approximation of MPC and the learned critic.
Following Appendix~\ref{app:bellman_residual} we can rollout this recursion into the explicit form
\begin{equation}
    \label{eq:blended_explicit}
    Q_0^{\lambda}
    =
    \hat{Q}_0
    + \lambda \left( \sum_{k=0}^{H-2} (\lambda \gamma)^k B_k
    + (\lambda \gamma)^{H-1} \bigl(l_{H-1} - \hat{Q}_{H-1}\bigr)\right),
\end{equation}
where $B_i = l_i + \gamma \hat{Q}_{i+1} - \hat{Q}_i$ is the one-step Bellman residual of the learned critic.
Setting $\lambda = 0$ recovers $\hat{Q}$ directly; $\lambda = 1$ recovers the standard MPC rollout; intermediate values let $\hat{Q}$ act as a global shaping term while preserving local model accuracy. In Section~\ref{par:lambda_study}, we additionally vary the blend weight of the critic at the end of the horizon separately, denoted $\lambda_f$; in all other experiments, $\lambda_f = \lambda$.

\subsection{Learning the Critic}
\label{sec:q_learning}
The critic $\hat Q_\theta$ is trained by stochastic gradient descent on the mean squared Bellman error~\cite{SuttonBarto}
\begin{equation}
    \mathcal{L}_Q(\theta) = \mathbb{E}_{(x_k,u_k,y_k)\sim\mathcal{D}}\left[\left(\hat Q_\theta(x_k,u_k)-y_k\right)^2\right].
\end{equation}
The target $y_k$ is constructed either from short rollout fragments or from the MPC continuation value $V^\lambda(x_{k+1}) = \min_u Q^\lambda(x_{k+1},u)$ which can be obtained for free after each MPC solve.

\paragraph{TD($n$) targets.}
A first option is to use standard $n$-step targets built from contiguous rollout fragments~\cite{SuttonBarto}. This is simple and efficient, and larger $n$ can stabilize the targets through the usual bias-variance trade-off. However, the method becomes sensitive to stale replay data when buffers are small or policy updates are infrequent.

\paragraph{MPC-based targets.}
A second option is to use this continuation value and form the one-step target $y_k^{\mathrm{MPC}} = l(x_k,u_k) + \gamma V^\lambda(x_{k+1})$, with the usual truncation at terminal transitions. In the offline variant, the planner stores the successor blended-Q-estimate during data collection and training reuses this cached value~\cite{Bhardwaj2020MPQ}. This is fast and benefits from the full planner lookahead, but the targets become stale as the critic changes. In the online variant, the successor value is recomputed during training by solving a fresh MPC problem for each minibatch element. This removes staleness and supports fully off-policy replay, but is typically prohibitively expensive: with iLQR, each batch element requires a converged optimal control solve, and shuffled minibatches prevent amortizing convergence across time steps.

\paragraph{Value ensembles.}
\label{par:value_ensembles}
Both target constructions remain bootstrapped and therefore inherit the usual risk of amplifying value-estimation errors~\cite{HasseltGS15}. In our setting this is especially problematic because MPC will immediately exploit critic mismatches induced by epistemic uncertainty. To reduce this effect, we follow the spirit of double Q-learning and ensemble bootstrapping~\cite{HasseltGS15,EnsembleBootstrapping2021} and use an ensemble critic with $N$ value heads. During each SGD step, one head is randomly selected for updating, while the remaining $N-1$ heads provide the target values. Their outputs can be aggregated by a (soft/hard) maximum, minimum, or mean operator; under our cost convention, the maximum is conservative, the minimum is optimistic, and we use the mean as a balanced compromise~\cite{EnsembleBootstrapping2021}. 

At inference time, the ensemble disagreement and the disagreement of the ensembles gradients also provide an uncertainty estimate of the learned critic. This gives us a way to downweigh the $\lambda$-contribution of in the MPC in regions where the Value-heads disagree. Contrary to~\cite{Lowrey2018}, where a softmax (softmin under our cost-convention) is used to encourage exploration, we compute the standard deviation across the normalized value-head outputs (and their gradients) and attenuate the learned-critic contribution by an exponential factor $\alpha = e^{-\kappa\sigma}$ where $\kappa$ is a tuning parameter.

\subsection{Learning the Dynamics Residual}
\label{sec:dynamics}

Many control tasks involve state variables that are directly observable, but for which no sufficiently accurate analytical dynamics model is available. In manipulation, for example, one may observe the pose and velocity of a grasped object without modeling the underlying contact dynamics explicitly. To allow the planner to reason over such quantities, we augment the state with components whose dynamics are incomplete and learn only the missing part from data.
Formally, we partition the state as $x = [x_p^\top, x_s^\top]^\top$, where $x_p \in \mathbb{R}^{n_p}$ denotes primary states with known dynamics and $x_s \in \mathbb{R}^{n_s}$ denotes secondary states whose evolution must be learned. We then use the composite dynamics
\begin{equation}
    \label{eq:f_tot}
    \dot{x} = f_\text{tot}(x, u) = f_p(x, u) + \mathbf{I}\, f_\phi(x, u),
\end{equation}
where $\mathbf{I} = \mathrm{diag}(m_1, \dots, m_n)$ with $m_i \in \{0,1\}$ is a binary mask selecting which state dimensions receive a learned residual correction, while dimensions with trusted dynamics, such as trivial integrators, are left unchanged.

We fit the residual from rollout data. Given samples $(x_k,u_k,x_{k+1})$, a one-step target is obtained from the finite difference $\dot{x}_k \approx (x_{k+1} - x_k)/\Delta t$; for states containing quaternion components, standard finite differences are replaced by logarithmic-map differences on $SO(3)$ so that the targets are well defined despite the $q \equiv -q$ sign ambiguity, as described in Appendix~\ref{app:quat}. Because one-step fitting alone does not control the error accumulated over repeated numerical integration inside the planner, we optimize both over mini-batches of the local regression loss and a rollout-consistency loss on contiguous trajectory fragments,
\begin{equation}
    \mathcal{L}(\phi) = \sum_k \bigl\|\mathbf{I}\bigl(f_\text{tot}(x_k,u_k) - \dot{x}_k\bigr)\bigr\|^2, \qquad \mathcal{L}_\text{roll}(\phi) = \frac{1}{R}\sum_{t=1}^{R} \Bigl\|\mathbf{I}(x_{k+t}^\text{pred} - x_{k+t}^\text{true})/\Delta t\Bigr\|^2,
\end{equation}
where $x_{k+t}^\text{pred}$ is obtained by recursively integrating $f_\text{tot}$ from the previous predicted state. This term penalizes compounding rollout error on the residual-enabled state components, rather than only minimizing local one-step errors. The resulting model allows MPC to plan directly in the augmented state space. Training then alternates between collecting rollouts, fitting the residual dynamics, and updating the learned critic. 

\subsection{Guiding the Critic}
\label{sec:learning_support}


\paragraph{Integration of Auxiliary Information.} Rather than learning directly from raw $(x,u)$, both learned modules in the solver have the option to tap into task-specific feature pre-processors,
\begin{equation}
    \dot{x} = f_p(x,u) + \mathbf{I}\,f_\phi\bigl(g_f(x,u,o)\bigr), \qquad
    \hat{Q}_\theta(x,u \mid o) = \hat{Q}_\theta\bigl(g_Q(x,u,o)\bigr),
\end{equation}
where $o$ denotes optional task context, such as a goal, object pose, or a latent visual embedding $z = e_\psi(o)$ produced by a shared visual encoder. The maps $g_f$ and $g_Q$ may differ and let the learned components operate in task coordinates rather than embodiment-specific raw states. For manipulation tasks, for instance, it can be convenient to compute inside $g_Q$ forward-kinematics-derived features such as $p_\mathrm{cube} - p_\mathrm{EE}$, $p_\mathrm{goal} - p_\mathrm{cube}$, and $v_\mathrm{EE}$ instead of raw joint variables.


\section{OCP Solver}
\label{sec:solver}

\subsection{iLQR}
\label{sec:ilqr}
Approaching the MPC from the MDP perspective (cf. Section~\ref{sec:prelim}), the DDP~\cite{MAYNE_DDP} based iLQR approach~\cite{Todorov2005AGI} is a natural choice for our optimal control solver. Given an initial control sequence $(u_i)_{i=0}^{H-1}$, a forward rollout under $x_{i+1}=f(x_i,u_i)$ produces a reference trajectory $(x_i,u_i)_{i=0}^{H-1}$ and its rollout value $V$. Following DDP, the backward pass uses the Bellman recursion $Q_i=l_i+\gamma Q_{i+1}$ and forms a local quadratic model of the action-value function at each time step,
\begin{equation}
    Q(\delta x,\delta u) \approx \frac{1}{2}
    \begin{bmatrix}
        1 \\
        \delta x \\
        \delta u
    \end{bmatrix}^{\!T}
    \begin{bmatrix}
        0 & Q_x^{T} & Q_u^{T} \\
        Q_x & Q_{xx} & Q_{xu} \\
        Q_u & Q_{ux} & Q_{uu}
    \end{bmatrix}
    \begin{bmatrix}
        1 \\
        \delta x \\
        \delta u
    \end{bmatrix}.
    \label{eq:ilqr_q_expansion}
\end{equation}
Minimizing this local model yields the affine feedback update
\begin{equation}
    \delta u_i = k_i + K_i\delta x_i, \qquad
    k_i = -Q_{uu,i}^{-1}Q_{u,i}, \qquad
    K_i = -Q_{uu,i}^{-1}Q_{ux,i}.
    \label{eq:ilqr_policy}
\end{equation}
iLQR drops the second-order dynamics tensor terms of full DDP, yielding the more economic Gauss-Newton approximation~\cite{Todorov2005AGI}. In the backward sweep, we compute the feed-forward terms $k_i$, feedback gains $K_i$, and the expected value-change terms used by the line search. The stage derivatives, dynamics Jacobians, and Hessians can be batch-computed before this sequential sweep, which allows us to GPU-parallelize the heavy computations related to auto-diff gradient and Hessian evaluation beforehand and keep the backward-sweep loop as lightweight as possible. We regularize $Q_{uu}$ using the scheme of Tassa et al.~\cite{Tassa2012}, with an aggressive schedule for decreasing or increasing the regularizer depending on line-search success.

The forward line search rolls out candidate controls $\hat u_i = u_i + \alpha k_i + K_i\,\mathrm{d}x(\hat x_i,x_i)$
from $\hat x_0=x_0$, where $\alpha$ is the line-search step size~\cite{Tassa2012}. The state displacement $\mathrm{d}x(\cdot,\cdot)$ is Euclidean for vector states and uses the manifold distance for quaternion states, as described in Appendix~\ref{app:quat}. We accept a candidate when the actual rollout improvement agrees with the expected local-model improvement, using either an Armijo condition or the ratio rule of Tassa et al.~\cite{Tassa2012}; otherwise $\alpha$ is reduced and the rollout is repeated.

\subsection{Augmented Lagrangian Method}
\label{sec:alm}

To handle equality and inequality constraints, we wrap the iLQR primal solve in an Augmented Lagrangian Method (ALM). Let $\tau=(x_0,u_0,\dots,x_{H-1},u_{H-1})$ denote the planned trajectory. We write the constrained finite-horizon problem with a generic additive objective $J(\tau)=\sum_{i=0}^{H-1}\tilde l_i(x_i,u_i)$, where $\tilde l_i$ will later be instantiated by the blended MPC stage objective:
\begin{equation}
\begin{aligned}
\min_{\tau}\quad & J(\tau)=\sum_{i=0}^{H-1}\tilde l_i(x_i,u_i) \\
\text{subject to}\quad & x_0=x, \\
& x_{i+1}=f(x_i,u_i), \\
& g_i(x_i,u_i)\le 0, \\
& h_i(x_i,u_i)=0, \qquad i=0,\dots,H-1 .
\end{aligned}
\label{eq:constrained_mpc}
\end{equation}
Let $c_i=[h_i^\top,g_i^\top]^\top$ collect constraint residuals and let $\mu_i=[\mu_{h,i}^\top,\mu_{g,i}^\top]^\top$ collect the corresponding multipliers. For fixed multipliers and penalties, the iLQR subproblem uses the augmented Lagrangian stage objective~\cite{Powell1969AMF,ALTRO},
\begin{equation}
    L_i^\mathrm{AL}(x_i,u_i,\mu_i,\rho_i)
    = \tilde l_i(x_i,u_i) + \mu_i^\top c_i(x_i,u_i) + \frac{1}{2}c_i(x_i,u_i)^\top I_{\rho,i}c_i(x_i,u_i).
    \label{eq:alm_stage_cost}
\end{equation}
The diagonal penalty matrix is $I_{\rho,i}=\mathrm{diag}(\rho_{h,i},m_i\odot\rho_{g,i})$, where $m_{ij}=0$ when $g_{ij}(x_i,u_i)<0$ and $\mu_{g,ij}=0$, and $m_{ij}=1$ otherwise. Thus, inactive inequality constraints do not contribute residuals, gradients, or Hessians to the local model.

ALM alternates between a primal and a dual step. In the primal step, iLQR minimizes $\sum_{i=0}^{H-1}L_i^\mathrm{AL}$ with $\mu_i$ and $I_{\rho,i}$ fixed, using the rollout and line-search procedure of Section~\ref{sec:ilqr}. After the inner solve returns the primal trajectory $(x_i^\star,u_i^\star)$, the dual step updates
\begin{equation}
    \mu_{h,i}^{+}=\mu_{h,i}+\rho_{h,i}h_i(x_i^\star,u_i^\star), \qquad
    \mu_{g,i}^{+}=\max\bigl(0,\mu_{g,i}+\rho_{g,i}g_i(x_i^\star,u_i^\star)\bigr).
    \label{eq:alm_dual_update}
\end{equation}
For constraints that remain violated, the penalty is increased as $\rho_i^{+}=\min(\Phi\rho_i,\rho_\mathrm{max})$ with $\Phi>1$. When the multiplier update vanishes, the residuals approach zero, and stationarity of the augmented subproblem reduces to the KKT stationarity conditions of the original constrained problem~\cite{Nocedal2006}.

\subsection{Blended-iLQR}
\label{sec:blendedCritic_objective_ilqr}

The constrained blended-MPC objective is obtained by substituting the ALM stage objective $L_i^\mathrm{AL}(x_i,u_i,\mu_i,\rho_i)$ into the blended recursion of Section~\ref{sec:blendedCritic_objective}. Define the marginalized continuation value $V_{i+1}^\lambda(x_{i+1})=\min_{u_{i+1}}Q_i^\lambda(x_{i+1},u_{i+1})$. For fixed ALM multipliers and penalties, the constrained recursion is
\begin{equation}
\begin{aligned}
Q_i^\lambda(x_i,u_i)
&=(1-\lambda)\hat Q(x_i,u_i)+\lambda\left(L_i^\mathrm{AL}(x_i,u_i,\mu_i,\rho_i)+\gamma V_{i+1}^\lambda(x_{i+1})\right) \\
&=\underbrace{\lambda L_i^\mathrm{AL}(x_i,u_i,\mu_i,\rho_i)+(1-\lambda)\hat Q(x_i,u_i)}_{\tilde L_i^\lambda(x_i,u_i)}+\lambda\gamma V_{i+1}^\lambda(f(x_i,u_i)).
\end{aligned}
\label{eq:constrained_mpc_recursion}
\end{equation}
At the last control step, the continuation is omitted and $Q_{H-1}^\lambda=\tilde L_{H-1}^\lambda$. Thus the blended AL-iLQR has the same dynamic-programming form as standard iLQR, but with blended ALM stage-objective $\tilde L_i^\lambda$ and effective discount $\lambda\gamma$. The backward pass therefore uses the same quadratic expansion as Eq.~\eqref{eq:ilqr_q_expansion}; only the derivatives of the local objective change. In particular, the constrained stage derivatives combine the ALM derivatives with the learned critic derivatives, e.g.,
\begin{equation}
    \tilde L_{x,i}^\lambda=\lambda(L_i^\mathrm{AL})_x+(1-\lambda)\hat Q_x, \qquad
    \tilde L_{u,i}^\lambda=\lambda(L_i^\mathrm{AL})_u+(1-\lambda)\hat Q_u,
\end{equation}
and analogously for $\tilde L_{xx,i}^\lambda$, $\tilde L_{uu,i}^\lambda$, and $\tilde L_{ux,i}^\lambda$. The full expansions are given in Appendix~\ref{app:q_expansion}. The remaining recursion is unchanged, except that the next-state value contribution is scaled by $\lambda\gamma$.

\section{Results}
\label{sec:experiments}

We evaluate the proposed framework in four complementary experiments. We begin with a runtime benchmark to verify our solver is amenable to the large-throughput requirement of reinforcement learning. We test the solver on randomly generated optimal-control problems to characterize how our implementation scales with batch size, hardware, and the method of derivative computation. Next, we isolate the effect of the critic blend on a point-mass task, where the optimal critic is available in closed form. We then study a quadrotor task with a biased dynamics model and an unmodeled suspended payload, which tests whether joint residual-dynamics learning and critic learning can compensate for model bias, capture missing state couplings, and remain compatible with quaternion kinematics. Finally, we consider Franka cube stacking as a contact-rich manipulation task to assess whether the framework can plan over incompletely modeled contact interactions and benefit from visual and haptic information for grasp detection.

\begin{figure}[t]
    \centering
    \includegraphics[width=1.0\linewidth]{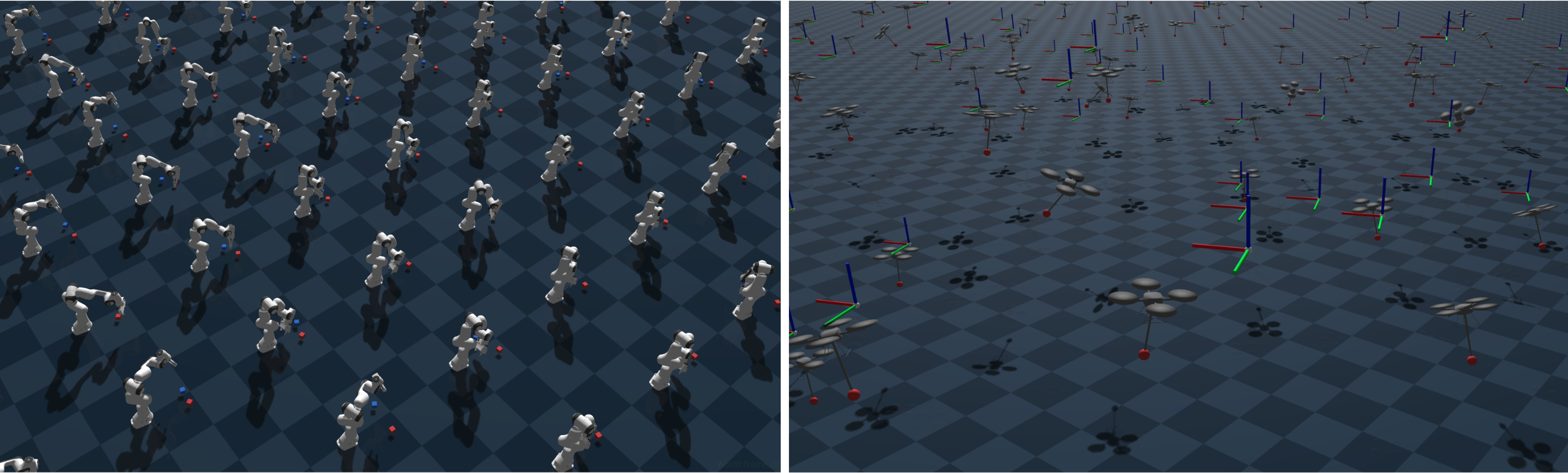}
    \caption{Parallel environments used for training and evaluation. Left: The stack-cube task with a 6-dof manipulator. The robot must pick up the target cube (blue) and stack it on top of the goal cube (red). Right: The quadrotor task, where the drone must hover at a target position while carrying an unmodeled, suspended payload. Both tasks are simulated using Genesis~\cite{Genesis}.}
    \label{fig:quadrotor_training}
\end{figure}

\paragraph{Speed Benchmarking.}
The experimental setup and detailed timing curves are given in Appendix~\ref{app:speed_benchmark}. The main result is that GPU execution becomes decisive once learned dynamics and critic models are part of the solve: on the tested hardware, batched GPU execution is roughly one to two orders of magnitude faster than CPU execution over the relevant batch-size range. In the purely non-learning setting, CPU execution is competitive only for the smallest batches, after which GPU execution overtakes it as batching exposes more parallel work. Figure~\ref{fig:runtime_scaling} further shows that the amortized runtime per problem decreases strongly up to about $10^4$ parallel QPs before saturating, and that automatic differentiation remains close to analytical derivatives in this benchmark.

\begin{figure}[t]
    \centering
    \includegraphics[width=0.85\linewidth]{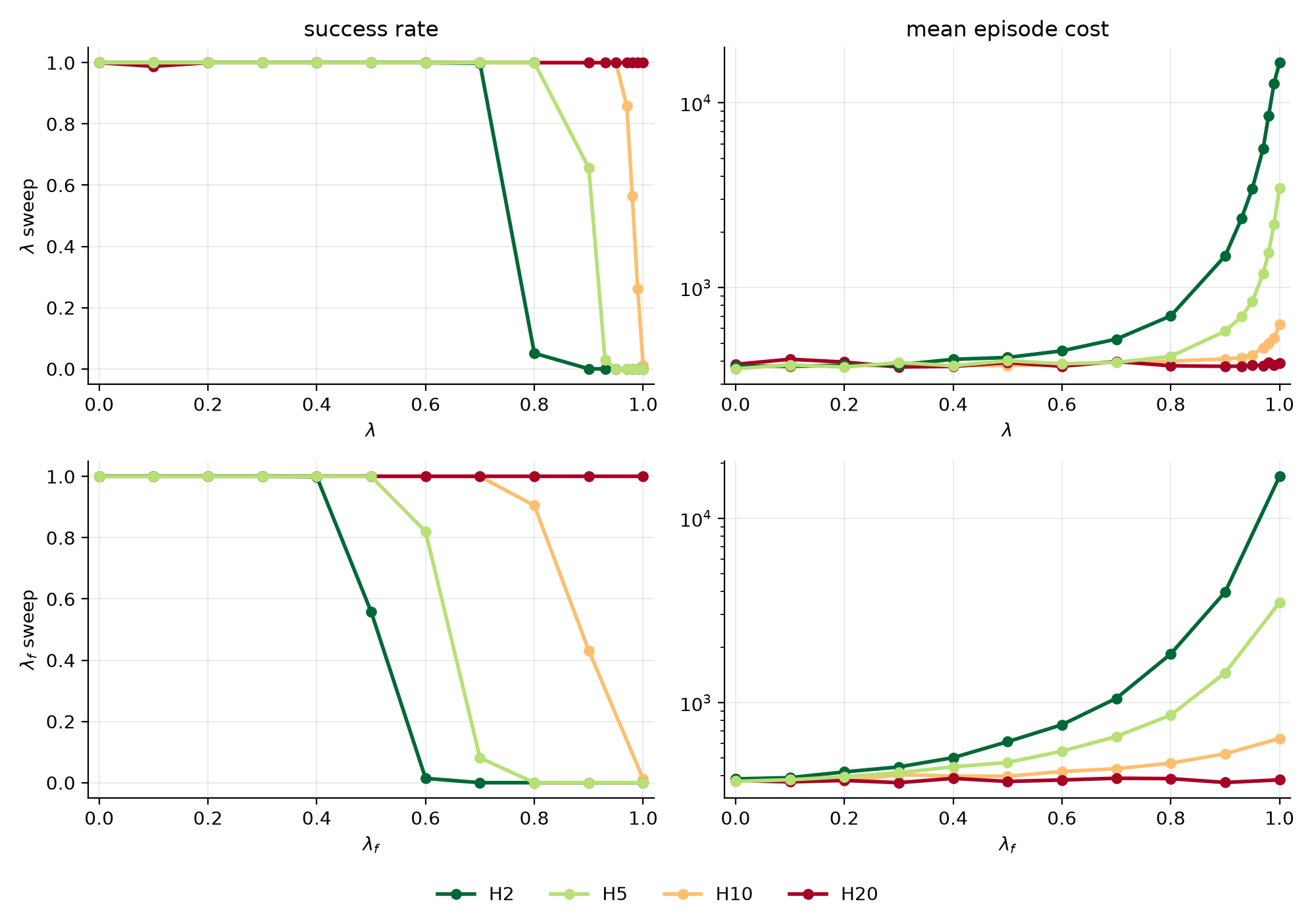}
    \caption{Point-mass reach task with the analytical LQR critic $Q^*, V^*$. Columns: success rate and mean episode cost. Top: all-stage blend, $\lambda = \lambda_f$ swept jointly. Bottom: terminal-only blend, $\lambda = 1$ and $\lambda_f$ swept.}
    \label{fig:pointmass_lambda}
\end{figure}

\paragraph{Isolating the Effect of the Critic Blend.}
\label{par:lambda_study}
To separate the effect of the blend from critic approximation errors, a force-controlled 2-D point mass must reach a fixed goal position and come to rest. For this linear-quadratic task, the optimal infinite-horizon discounted $Q^*$ and $V^*$ follow in closed form from the discrete algebraic Riccati equation and take the place of the learned critic (Appendix~\ref{app:pointmass}). We sweep the blend for horizons $H \in \{2,5,10,20\}$ in two settings: blending $Q^*$ at all stages with $\lambda_f = \lambda$, and blending only the terminal value with $\lambda_f$, while $\lambda = 1$ at all other stages. Figure~\ref{fig:pointmass_lambda} shows the results. Without the critic ($\lambda = \lambda_f = 1$), MPC overshoots the goal for $H \in \{2,5,10\}$ and fails the task, whereas $H = 20$ succeeds. Lowering $\lambda_f$ towards zero recovers the success rate and cost of the optimal policy at all horizons. Blending at all stages reaches the same performance already at values of $\lambda$ closer to $1$. We interpret this as the stage-wise critic terms supplying long-range information along the whole planned trajectory, in addition to the terminal value. These results rely on the exact $Q^*$; Section~\ref{sec:limitations} discusses why learned critics restrict $\lambda$ to values close to $1$.

\begin{figure}[h]
    \centering
    \includegraphics[width=1.0\linewidth]{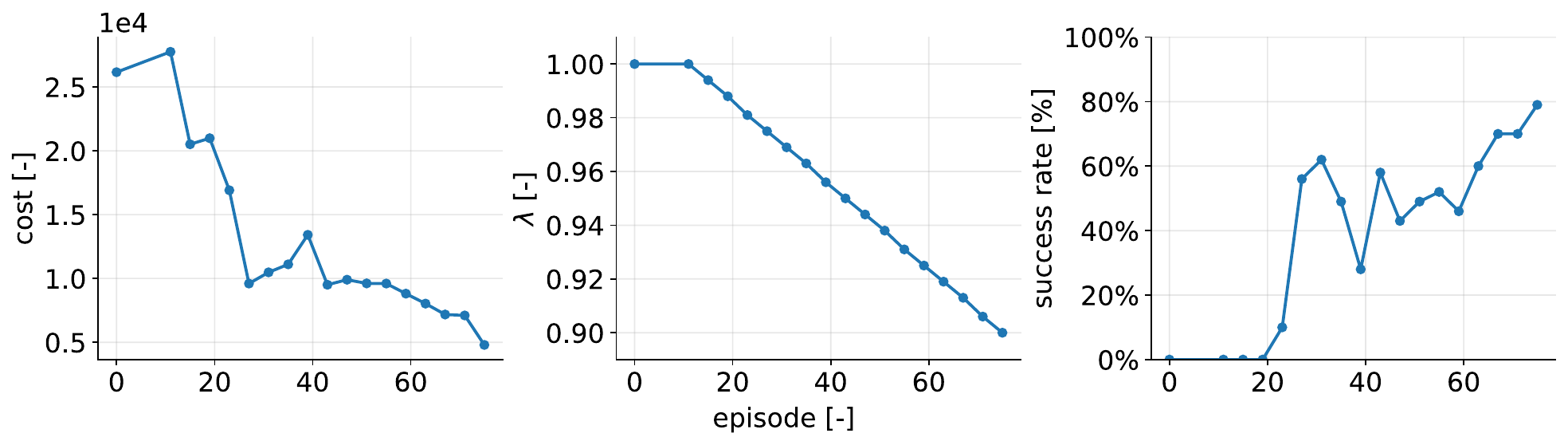}
    \caption{Quadrotor learning progress while decreasing the critic blend parameter $\lambda$. Early updates behave like nominal MPC and fail under the biased incomplete model (cf. Section~\ref{sec:blendedCritic_objective}); once the learned critic is allowed to shape the objective more strongly, the success rate increases and the cost drops.}
    \label{fig:quadrotor_lambda}
\end{figure}

\paragraph{Overcoming Biased Dynamics and Unmodeled State-Dependencies.}
\label{sec:quadrotor_experiment}
In this experiment, a quadrotor starts in a random pose and must hover to a target position while carrying an unmodeled payload suspended by a rod. With all learning components disabled, the nominal MPC baseline reaches a success rate of only $0.05$ and a mean cost of $2.6e4$, with most failures caused by crashes. The learned critic variant, which combines residual-dynamics learning with the blended critic objective, raises the success rate to $0.79$ and lowers the mean cost to $4.7e3$. As an adaptive-control baseline, an idealized incremental nonlinear dynamic inversion (INDI) controller~\cite{Smeur2016INDI} without sensor noise or delay reaches a higher success rate of $0.94$, but a higher mean cost of $5.6e3$, since it overshoots more and settles more slowly around the goal (Appendix~\ref{app:quadrotor_experiment_setup}). In our setting, learning the dynamics model alone is not sufficient to solve the task reliably. We attribute this to the short look-ahead of $1$s for the MPC, not capturing the fatal swing-up motion of the pendulum early enough, and the approximation bias of the learned dynamics model, requiring careful alignment of planning frequency, integration scheme, and look-ahead distance to trade off approximation bias with real-time-ness. However, we observe that combining the learned dynamics model with the learned critic, which injects infinite-horizon MDP information into the MPC, yields more conservative approaches towards the goal, keeping the oscillations bounded and avoiding crashing. The learned critic can drive the system away from high action-value regions that lead to failure, including the ones with high pendulum deflection.
This observation is broadly in line with the remarkable success of model-free policy-gradient methods, which achieve high task performance not by meticulously modeling the underlying dynamics but by directly learning mitigation strategies that avoid task failure~\cite{Peters2010_policy_gradient}.
To further demonstrate the sample-efficiency claim, we establish a PPO baseline~\cite{schwarke2025rslrl} with the hyperparameters of Table~\ref{tab:ppo_settings} reaching comparable asymptotic performance but requiring training for 1.5 hours on $1e4$ parallel environments, collecting roughly $1e8$ environment transitions (collected at 200 Hz, the policy frequency), whereas our learned-critic achieves the reported performance after roughly 30 minutes using $10^5$ transitions (collected at 20 Hz, the planner frequency, cf. Appendix~\ref{app:solver_and_learning_settings}). Figure~\ref{fig:quadrotor_lambda} shows how this improvement emerges during training: as $\lambda$ is decreased, the learned critic gradually takes over a larger share of the objective, correcting the nominal behavior and improving both success rate and cost. Detailed task and training settings are given in Appendix~\ref{app:quadrotor_experiment_setup}, whereas empirical evidence for the efficacy of our approach and further insights into the learned critic are given in Appendix~\ref{app:quadrotor_experiment_insights}.

\paragraph{Learning to Plan over Contacts Using Visual and Haptic Information.}
\label{sec:cube_stacking_experiment}
In this experiment, a 6-DOF manipulator must pick up a cube and stack it precisely on top of another cube using only a simple kinematic robot model and no explicit contact model.
Unsurprisingly, a naive grasp-uninformed nominal MPC with a single global cost function fails completely (success rate of $0$, stuck in pre-grasp pose) since it cannot reason over the grasp. Following Appendix~\ref{app:stacking_experiment_setup}, we add to the state vector the cube's position and a grasp-score, which is initialized from an external grasp detection using tactile sensors, and replace the global cost with a multi-phase cost function. This modification raises the nominal baseline to $0.361$ success rate. Further learning a residual to the grasp-score state lifts the success rate to moderate $0.401$. Employing a learned critic, which is conditioned on visual depth and haptic observations (cf. Figure~\ref{fig:depth}) improves task performance significantly to $0.94$ success rate. We observe more delicate and better-informed recovery behaviors than with nominal MPC, including corrections after failed grasp attempts and poor approaches that tip over the target cube. More insights into the task setup and further discussions of the results are given in Appendix~\ref{app:stacking_experiment_setup} and Appendix~\ref{app:stacking_experiment_insights}.

\section{Limitations}
\label{sec:limitations}

A first limitation is that the solver relies on first- and second-order derivatives of the learned residual and critic models, and optimization remains sensitive to irregular derivatives. This restricts the blend: with the exact critic of Section~\ref{par:lambda_study}, lower values of $\lambda$ consistently recover the performance of the optimal policy, whereas with learned critics only modest reductions from $\lambda=1$, roughly in the range $0.98$ to $0.9$, outperformed the nominal baselines in our experiments. Since the weight of the nominal rollout decays as $(\gamma\lambda)^k$ along the horizon (Eq.~\eqref{eq:blended_explicit}), low $\lambda$ removes the model information that would compensate imperfect critic derivatives, particularly at long horizons (cf. Appendix~\ref{app:stacking_experiment_insights}). We therefore start training at $\lambda=1$, lower $\lambda$ only once the critic has acquired sufficient state-space coverage, and mitigate irregular derivatives by disagreement-based attenuation of the critic, smoothing, and clipping. Nevertheless, poor gradients can still steer the optimizer toward uninformative regions of the state space, degrade data collection early-on, and ultimately stall training; stronger training-time regularization, efficient off-policy target construction and priority-sampling from the replay-buffer should therefore be investigated. A second limitation concerns real-time deployment on a single system. While all solver computations are batched across environments, and costs and derivatives are evaluated in batch over all time steps, the forward rollout, the Riccati recursion, and the line search remain sequential over the horizon, which precludes replanning at rates above $100\,\mathrm{Hz}$ in the current implementation. However, iLQR returns the time-indexed local feedback policy $\hat u_i = u_i + K_i\,\mathrm{d}x(\hat x_i, x_i)$ around the planned trajectory, so the planner can run at a low rate while this policy closes the loop at the control rate.

\section{Conclusion}
\label{sec:conclusion}
To conclude, this paper presented a learning-augmented constrained MPC framework for planning under model mismatch. While dynamics learning enabled the planner to account for directly observable but incompletely modeled state variables, the main driver of the observed performance gains was the integration of a learned critic into the gradient-based optimization framework. By providing corrective descent directions during the backward recursion, the critic injected long-horizon behavioral information that is inaccessible to a finite-horizon MPC objective alone. In this way, the method mitigated the inherent myopia of MPC by leveraging value information from the underlying infinite-horizon MDP. Across the reported runtime and control experiments, the framework scaled to thousands of parallel optimal control problems and improved performance under biased and incomplete dynamics models. Future work will focus on addressing the aforementioned limitations and on stronger critic pre-training, including representation learning from video demonstrations.


\clearpage
\appendix

\section{Bellman-Residual Form of the Blended Objective}
\label{app:bellman_residual}
We have H control steps, hence the planned rollout contains the state-action pairs $(x_i, u_i)_{i=0}^{H-1}$ and terminal state $x_H$. For compactness, write $Q_i = Q(x_i, u_i)$, $\hat Q_i = \hat Q(x_i, u_i)$, and $l_i = l(x_i, u_i)$. 

For $i = 0, \dots, H-2$, the blended recursion in the main text reads
\begin{equation*}
Q_i^{\lambda} = (1-\lambda)\hat Q_i + \lambda\left(l_i + \gamma Q_{i+1}^{\lambda}\right),
\end{equation*}
where $\hat Q_i$ is the learned Q-estimate and $Q_i^{\lambda}$ is the MPC-blended Q estimate at timestep $i$. At the last control step we close the recursion with
\begin{equation*}
Q_{H-1}^{\lambda} = (1-\lambda)\hat Q_{H-1} + \lambda l_{H-1}.
\end{equation*}
To expose the relation to temporal-difference style corrections, define the deviation of the blended objective from the learned critic by
\begin{equation*}
\delta_i = Q_i^{\lambda} - \hat Q_i.
\end{equation*}
Subtracting $\hat Q_i$ from both sides of the recursion gives
\begin{equation*}
\delta_i
=
(1-\lambda)\hat Q_i + \lambda\left(l_i + \gamma Q_{i+1}^{\lambda}\right) - \hat Q_i
=
\lambda\left(l_i + \gamma Q_{i+1}^{\lambda} - \hat Q_i\right).
\end{equation*}
Now add and subtract $\gamma \hat Q_{i+1}$ inside the parentheses:
\begin{equation*}
\delta_i
=
\lambda\left(l_i + \gamma \hat Q_{i+1} - \hat Q_i + \gamma (Q_{i+1}^{\lambda} - \hat Q_{i+1})\right).
\end{equation*}
This separates the local Bellman residual of the learned critic from the downstream blended correction. Define the one-step Bellman residual
\begin{equation*}
B_i = l_i + \gamma \hat Q_{i+1} - \hat Q_i.
\end{equation*}
Using $\delta_{i+1} = Q_{i+1}^{\lambda} - \hat Q_{i+1}$, we obtain the recursion
\begin{equation*}
\delta_i = \lambda (B_i + \gamma \delta_{i+1}),
\end{equation*}
At the last control step, the terminal approximation gives
\begin{equation*}
\delta_{H-1} = Q_{H-1}^{\lambda} - \hat Q_{H-1} = \lambda\left(l_{H-1} - \hat Q_{H-1}\right).
\end{equation*}

For any intermediate step we get
\begin{equation*}
\delta_i = \lambda \left( \sum_{k=0}^{H-i-2} (\lambda \gamma)^k B_{i+k} + (\lambda \gamma)^{H-i-1} \delta_{H-1} \right).
\end{equation*}
Setting $i=0$ and substituting back $Q_0^{\lambda} = \hat Q_0 + \delta_0$ yields
\begin{equation*}
Q_0^{\lambda}
=
\hat Q_0
+ \lambda \left( \sum_{k=0}^{H-2} (\lambda \gamma)^k B_k
+ (\lambda \gamma)^{H-1} \left(l_{H-1} - \hat Q_{H-1}\right) \right).
\end{equation*}
Hence, the blended objective can be interpreted as the learned critic plus a geometrically weighted accumulation of local Bellman-residual corrections along the planned trajectory. In particular, when the learned critic satisfies the Bellman relation well along the planned rollout, the residual terms $B_i$ are small and the blended objective stays close to $\hat Q_0$; when the local model-based rollout disagrees with the learned critic, the residual terms accumulate and correct the objective accordingly.

\section{Quadratic Q Expansion for blended ALM-iLQR}
\label{app:q_expansion}
For the ALM-iLQR the gradients and hessians for the quadratic Q expansion in each timestep read:
\begin{equation*}
\begin{aligned}
Q_x &= \widetilde{L}_x + f_x^{\top} V_x^{\prime},\\
Q_u &= \widetilde{L}_u + f_u^{\top} V_u^{\prime},\\
Q_{xx} &= \widetilde{L}_{xx} + f_x^{\top} V_{xx}^{\prime} f_x + V_x^{\prime} \cdot f_{xx},\\
Q_{uu} &= \widetilde{L}_{uu} + f_u^{\top} V_{uu}^{\prime} f_u + V_x^{\prime} \cdot f_{uu},\\
Q_{ux} &= \widetilde{L}_{ux} + f_u^{\top} V_{xx}^{\prime} f_x + V_x^{\prime} \cdot f_{ux}.
\end{aligned}
\end{equation*}
with the learned Q-function terms, and using a locally-linear approximation of the constraint function $c_i \approx c_i^0 + c_{x,i}\delta x_i + c_{u,i}\delta u_i$,
\begin{equation*}
\begin{aligned}
    \widetilde{L}_x &= \lambda (L_i^\mathrm{AL})_x + (1-\lambda) \hat Q_x,
    & \qquad (L_i^\mathrm{AL})_x &= l_x + c_{x,i}^\top(\mu_i + I_{\rho,i} c_i),\\
    \widetilde{L}_u &= \lambda (L_i^\mathrm{AL})_u + (1-\lambda) \hat Q_u,
    & \qquad (L_i^\mathrm{AL})_u &= l_u + c_{u,i}^\top(\mu_i + I_{\rho,i} c_i),\\
    \widetilde{L}_{xx} &= \lambda (L_i^\mathrm{AL})_{xx} + (1-\lambda) \hat Q_{xx},
    & \qquad (L_i^\mathrm{AL})_{xx} &\approx l_{xx} + c_{x,i}^\top I_{\rho,i} c_{x,i},\\
    \widetilde{L}_{uu} &= \lambda (L_i^\mathrm{AL})_{uu} + (1-\lambda) \hat Q_{uu},
    & \qquad (L_i^\mathrm{AL})_{uu} &\approx l_{uu} + c_{u,i}^\top I_{\rho,i} c_{u,i},\\
    \widetilde{L}_{ux} &= \lambda (L_i^\mathrm{AL})_{ux} + (1-\lambda) \hat Q_{ux},
    & \qquad (L_i^\mathrm{AL})_{ux} &\approx l_{ux} + c_{u,i}^\top I_{\rho,i} c_{x,i}.
\end{aligned}
\end{equation*}

the next-state value-function gradient $V_x'$ and hessian $V_{xx}'$ are carried over in the recursion.

\section{Runtime Scaling Benchmark}
\label{app:speed_benchmark}
To isolate solver throughput, we benchmark the batched iLQR implementation on random control instances with an 8-dimensional state, a 4-dimensional action, a planning horizon of 32 steps, and a planning time step of $0.05\,\mathrm{s}$. The learned dynamics model and learned critic are fully connected MLPs with three hidden layers of width 256 and ELU activations. Each timing run measures a single iLQR iteration, comprising one forward rollout, one backward pass, up to six regularization trials, and up to eight line-search evaluations. We discard two warmup runs and report averages over 20 measured runs. The benchmark is executed on an NVIDIA RTX 5080 mobile GPU (175 W) and an AMD Ryzen 9 9955HX CPU with 64 GB RAM.

\begin{figure}[h]
    \centering
    \includegraphics[width=\linewidth]{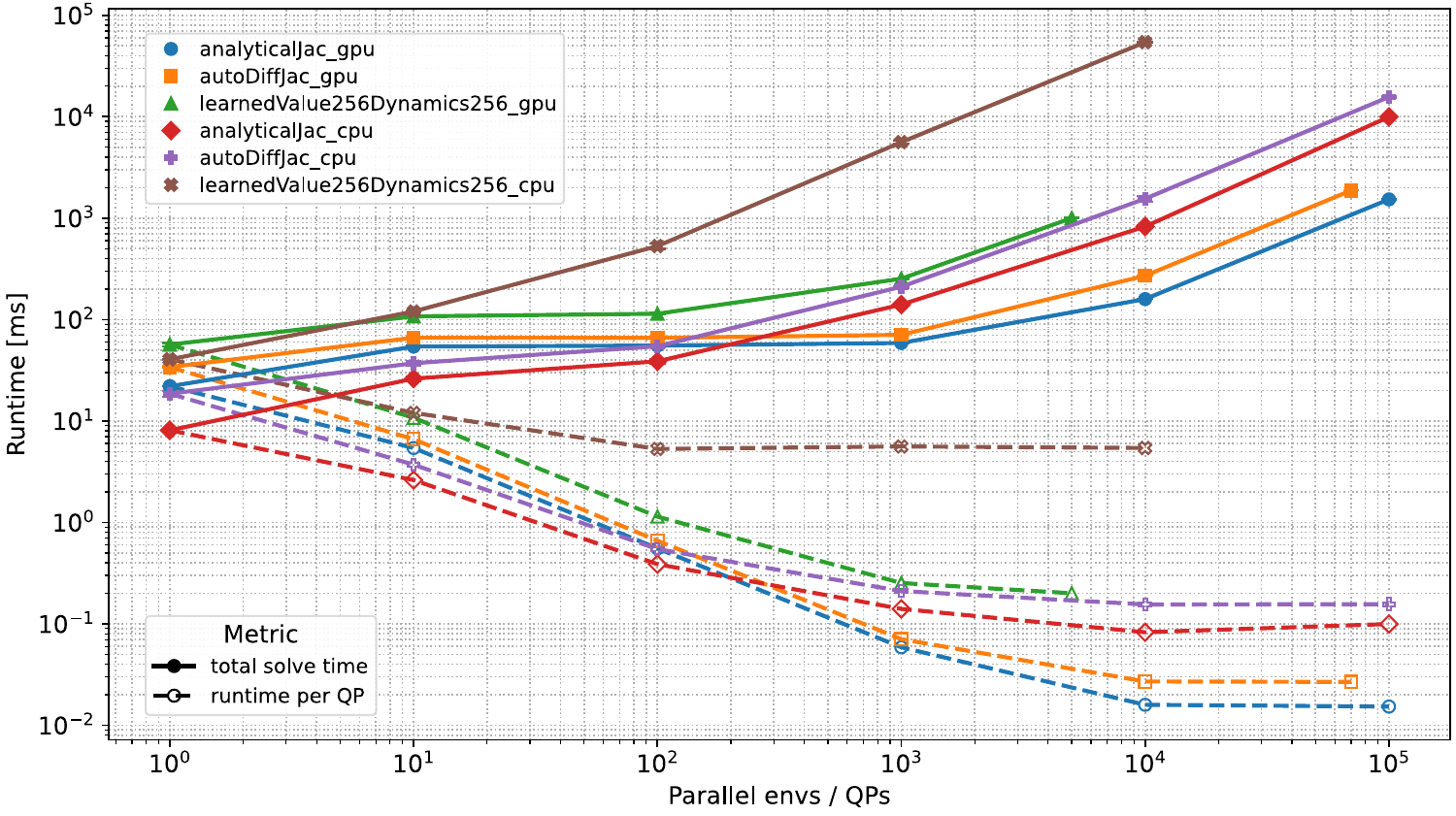}
    \caption{Runtime scaling of one batched iLQR iteration with analytical and autodiff derivatives on CPU and GPU. GPU execution becomes decisive once learned models are part of the solve, while autodiff remains close to analytical derivatives; runtime per QP decreases with batching up to about $10^4$ parallel problems before flattening on the tested hardware.}
    \label{fig:runtime_scaling}
\end{figure}

The figure separates total solve time from amortized runtime per QP. For the learned-model setting, the GPU curves stay well below the CPU curves across the relevant batch sizes, yielding roughly one to two orders of magnitude lower runtime on the tested hardware. For the analytical setting, CPU execution is competitive only at the smallest batches, after which GPU execution overtakes it as batching exposes more parallel work. The amortized runtime decreases strongly up to about $10^4$ parallel QPs and then saturates, which in our setup coincides with the practical hardware limit of the tested mobile GPU.

\section{Point-Mass Experiment Details}
\label{app:pointmass}
The point mass has mass $m = 2\,\mathrm{kg}$, state $x = [p^\top, v^\top]^\top \in \mathbb{R}^4$ with planar position $p$ and velocity $v$, and is driven by a planar force $u \in [-20, 20]^2\,\mathrm{N}$ with $\dot p = v$ and $\dot v = u/m$. The stage cost is $\ell(x,u) = \beta_p \lVert p - p^\star \rVert_2^2 + \beta_v \lVert v \rVert_2^2 + \beta_u \lVert u \rVert_2^2$ with $\beta_p = 1$, $\beta_v = 0.05$, and $\beta_u = 0.1$. Linearizing the discretized dynamics at $\Delta t_{\mathrm{plan}} = 0.1\,\mathrm{s}$ gives $(A, B)$, and the discounted algebraic Riccati equation with $\gamma = 0.95$ gives $P$, so that $V^*(x) = e^\top P e$ and $Q^*(x,u) = \ell(x,u) + \gamma\, V^*(A e + B u)$ in the error coordinates $e = x - [p^{\star\top}, 0]^\top$. An episode of $10\,\mathrm{s}$ counts as successful once the position error falls below $0.1\,\mathrm{m}$ and the speed below $0.05\,\mathrm{m/s}$. Each sweep runs $10^3$ parallel environments and holds each blend value for one episode, $\lambda \in \{1, 0.99, 0.98, 0.97, 0.95, 0.93, 0.9, 0.8, \dots, 0\}$ for the all-stage sweep and $\lambda_f \in \{1, 0.9, \dots, 0\}$ for the terminal-only sweep.

\section{Quadrotor Experiment Details}
\label{app:quadrotor}

\subsection{Task Setup}
\label{app:quadrotor_experiment_setup}  

\paragraph{Dynamics Model.}
The nominal MPC model uses the rigid-body state and control given by
\begin{equation*}
\begin{aligned}
x &= \begin{bmatrix}
p_{WB}^W \\
q_{WB} \\
v_{WB}^W \\
\omega_{WB}^B
\end{bmatrix},
\qquad
u = \begin{bmatrix}
a_z^B \\
\alpha_{WB}^B
\end{bmatrix},
\end{aligned}
\end{equation*}
The virtual controls are subject to componentwise box constraints,
\begin{equation*}
u_{\min} \leq u \leq u_{\max},
\qquad
a_z^B \in [-20,40]\,\mathrm{m}\,\mathrm{s}^{-2},
\qquad
\alpha_{WB}^B \in [-40,40]^3\,\mathrm{rad}\,\mathrm{s}^{-2},
\end{equation*}
equivalently written in the inequality-constraint form as $h_{\min}(u)=u_{\min}-u\leq 0$ and $h_{\max}(u)=u-u_{\max}\leq 0$.
For the quaternion block, the solver uses the reduced local coordinates and projection matrix
\begin{equation*}
    x_\mathrm{red}=
    \begin{bmatrix}p_{WB}^W\\\phi_{WB}\\v_{WB}^W\\\omega_{WB}^B\end{bmatrix}, \qquad
    E=
    \begin{bmatrix}
        I & 0 & 0 & 0 \\
        0 & G(q_{WB}) & 0 & 0 \\
        0 & 0 & I & 0 \\
        0 & 0 & 0 & I
    \end{bmatrix},
\end{equation*}
so the quaternion state is optimized in a three-dimensional tangent coordinate $\phi_{WB}\in\mathbb{R}^3$ rather than directly in $q_{WB}\in S^3$.
The dynamics are
\begin{equation*}
\dot x = f(x,u) = \begin{bmatrix}
v_{WB}^W \\
	\frac{1}{2}\, q_{WB} \otimes \begin{bmatrix}
0 \\
\omega_{WB}^B
\end{bmatrix} \\
g^W + R(q_{WB}) \begin{bmatrix}
0 \\
0 \\
a_z^B
\end{bmatrix} \\
\alpha_{WB}^B
\end{bmatrix}.
\end{equation*}
where $g^W = \begin{bmatrix} 0 \\ 0 \\ -g \end{bmatrix}$ and $R(q_{WB})$ maps body-frame accelerations into the world frame. To inject structured model mismatch, the simulator additionally exposes a swing-load state with roll- and pitch pendulum coordinates,
\begin{equation*}
x^{\mathrm{aug}} = \begin{bmatrix}
x \\
\theta \\
\dot \theta
\end{bmatrix}, \qquad \theta \in \mathbb{R}^2,
\end{equation*}
corresponding to a $0.172\,\mathrm{kg}$ suspended load ($0.15\,\mathrm{kg}$ payload, $0.02\,\mathrm{kg}$ rod, $0.002\,\mathrm{kg}$ bearing) attached to a $0.15\,\mathrm{m}$ rod. These state variables are observed during training but are not modeled in the nominal rigid-body dynamics above; their effect on the translational and rotational motion must therefore be captured from data alone.

The model bias $\bar m = m / 1.5$ and $\bar J_B = 1.5 J_B$ is introduced in the the actuator model
\begin{equation*}
\begin{aligned}
\dot v_{WB}^W &= g^W + \bar m^{-1} R(q_{WB}) f_B(\omega^2), \\
\dot \omega_{WB}^B &= \bar J_B^{-1}\!\left(\tau_B(\omega^2) - \omega_{WB}^B \times \bar J_B \omega_{WB}^B\right),
\end{aligned}
\end{equation*}

which computes the the squared rotor-speed commands $\omega_i^2$ from the virtual controls $u$ using the wrench model
\begin{equation*}
\begin{aligned}
f_B(\omega^2) &= \left[\begin{smallmatrix}
0 \\
0 \\
k_f \sum_{j=1}^{4} \omega_j^2
\end{smallmatrix}\right], \qquad
	\tau_B(\omega^2) = \sum_{j=1}^{4} r_j \times \left[\begin{smallmatrix}
0 \\
0 \\
k_f \omega_j^2
\end{smallmatrix}\right] + \left[\begin{smallmatrix}
0 \\
0 \\
k_d \sum_{j=1}^{4} s_j \omega_j^2
\end{smallmatrix}\right],
\end{aligned}
\end{equation*}

with $k_f$ and $k_d$ the force and drag coefficients, $r_j$ the rotor-arm vectors, and $s_j \in \{-1,+1\}$ the propeller spin directions.  Hence the benchmark combines parameter bias on the modeled rigid-body states with additional unmodeled state dependencies from the swing-load coordinates.

\paragraph{Cost Function}
The quadrotor stage cost is a quadratic hover-tracking objective, where $(\cdot)^\star$ denotes the hover goal state,
\begin{equation*}
\begin{aligned}
\ell(x,u) = \tfrac{1}{2} \Bigl(
&\beta_p \lVert p_{WB}^W - p_{WB}^{W,\star} \rVert_2^2
+ \beta_q \lVert \operatorname{rotvec}(q_{WB}^\star \otimes q_{WB}^{-1}) \rVert_2^2 \\
&+ \beta_v \lVert v_{WB}^W - v_{WB}^{W,\star} \rVert_2^2
+ \beta_\omega \lVert \omega_{WB}^B - \omega_{WB}^{B,\star} \rVert_2^2 \\
&+ \beta_u \lVert u \rVert_2^2
\Bigr),
\end{aligned}
\end{equation*}
where $\beta_p$, $\beta_q$, $\beta_v$, $\beta_\omega$, and $\beta_u$ are positive cost weights.

\paragraph{INDI Baseline.}
INDI~\cite{Smeur2016INDI} computes control increments from measured accelerations and thereby compensates unmodeled forces and moments, including those of the suspended payload, without an explicit model of them. We evaluate an idealized variant, in which these measurements are free of noise and delay, on the same task and with the same cost function as the MPC variants.

\subsection{Experimental Insights}
\label{app:quadrotor_experiment_insights}

\begin{figure*}[t]
    \centering
    \includegraphics[width=\linewidth]{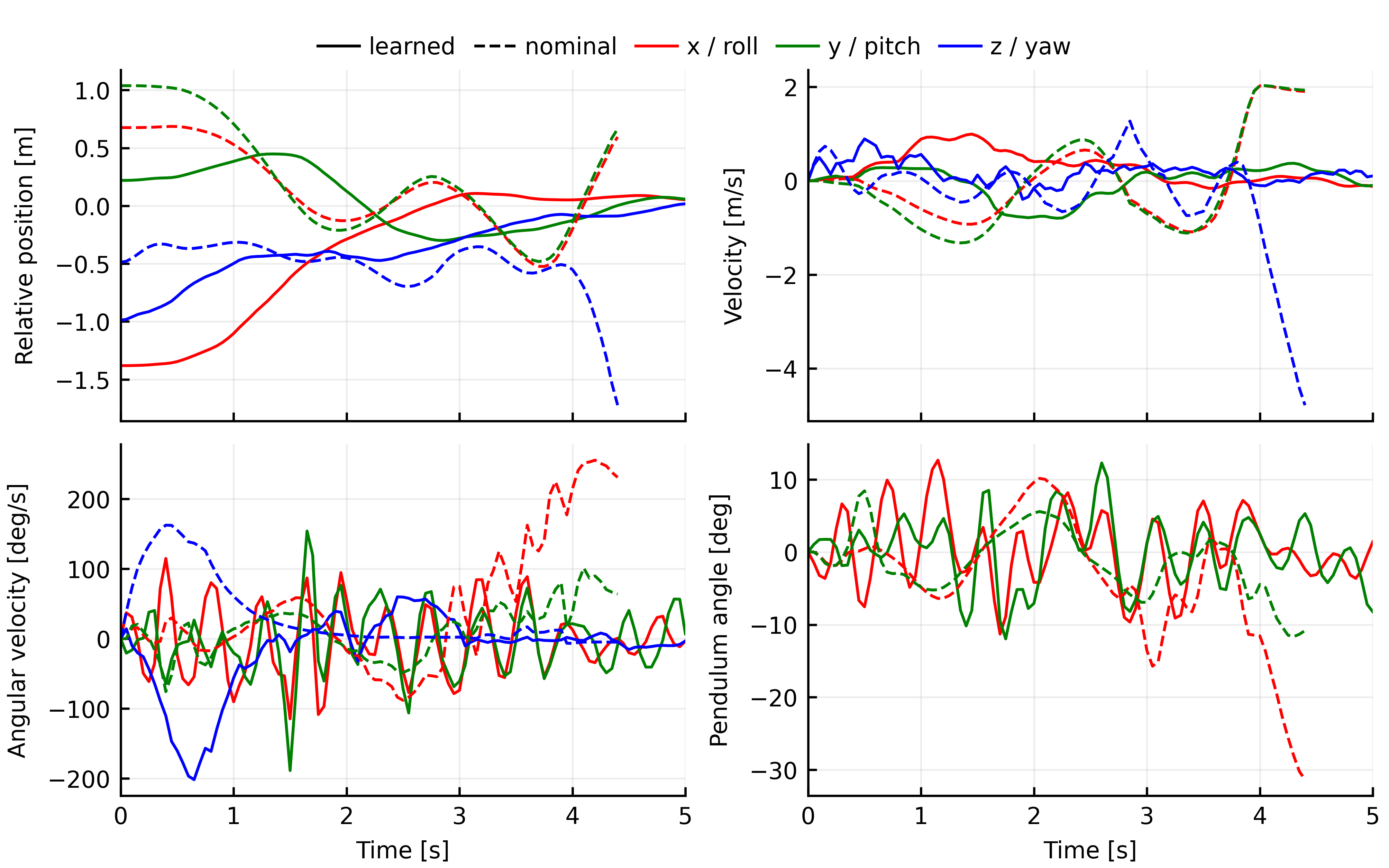}
    \caption{Two closed-loop trajectories under the biased incomplete quadrotor model. Solid lines denote the trajectories stabilized by the learned critic controller; dashed lines denote the nominal MPC. Different random initial states are shown for the two controllers. The nominal controller initially moves toward the goal but excites the suspended load, which then lead to tip-over late into the flight. The learned controller approaches the goal more conservatively, keeping the pendulum motion bounded and stabilizing the system throughout the episode.}
    \label{fig:quadrotor_rollout_compare}
\end{figure*}

\paragraph{Failure mode of nominal MPC.}
Figure~\ref{fig:quadrotor_rollout_compare} compares representative closed-loop trajectories of nominal MPC and the blended-critic controller under the biased incomplete model (random seeds). The nominal controller initially reduces the position error, but it does so with the wrong inertial model and without explicit payload dynamics. This injects excessive rotational motion into the suspended load, which amplifies and destabilizes the quadrotor. Around the end of the episode, the nominal controller tips over, loses altitude and crashes. The learned critic avoids this failure: by learning to favor a lower-velocity approach, it keeps the pendulum motion bounded, avoiding catastrophic tip-over even under presence of severe cross-couplings.



\begin{figure*}[t]
    \centering
    \begin{minipage}[t]{0.48\textwidth}
        \centering
        Blended Critic MPC\\[2pt]
        \includegraphics[width=\linewidth]{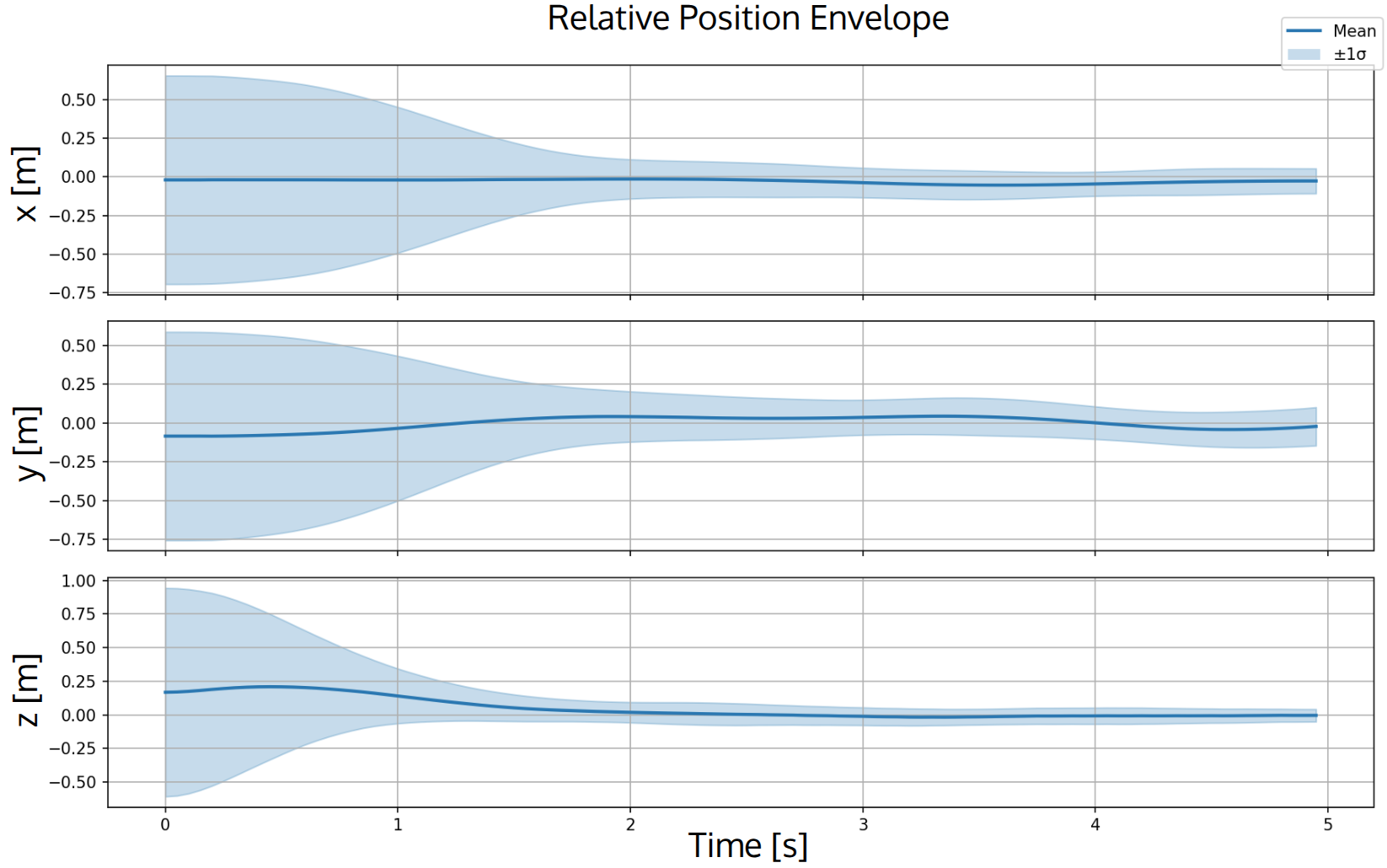}
        \includegraphics[width=\linewidth]{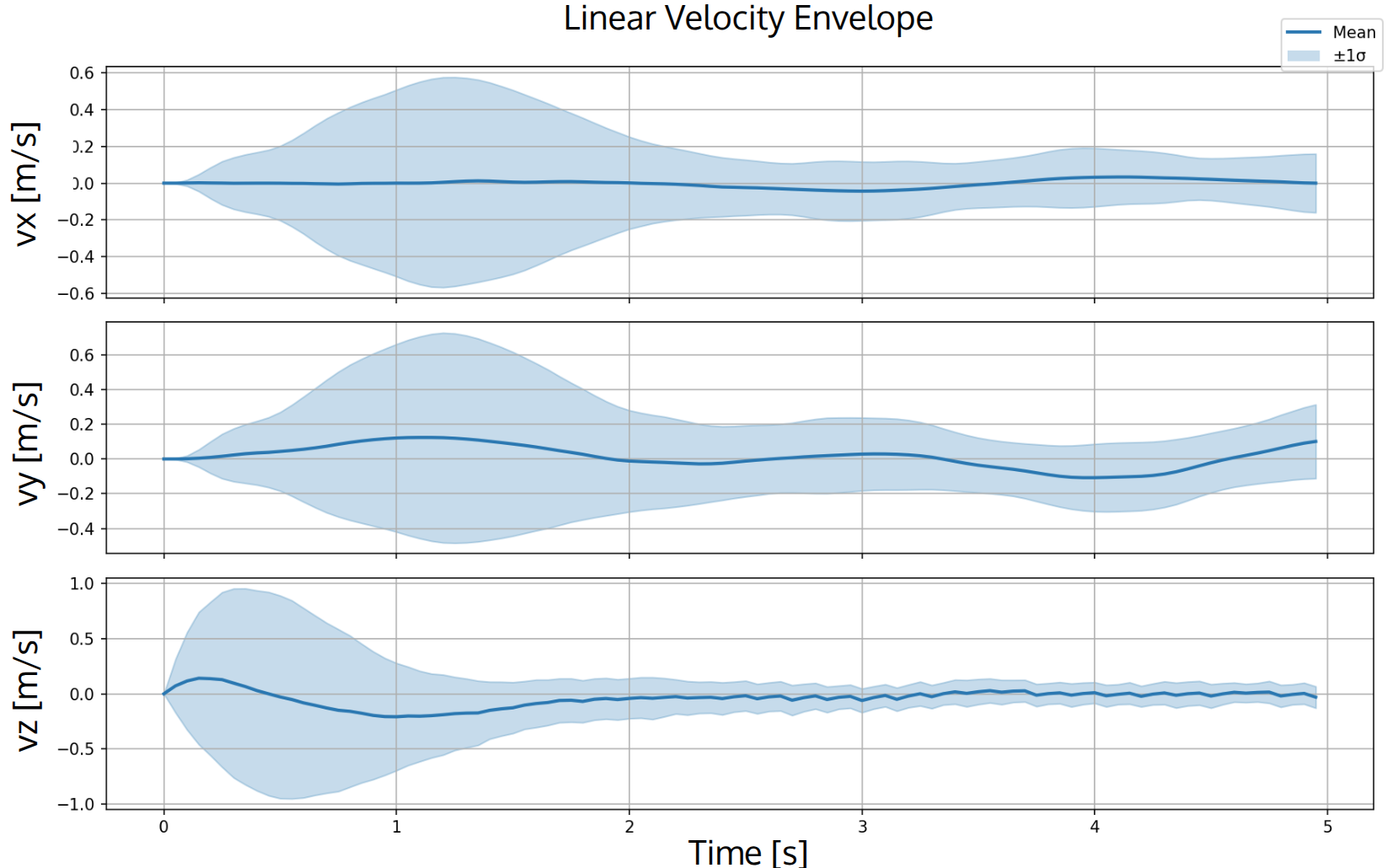}
        \includegraphics[width=\linewidth]{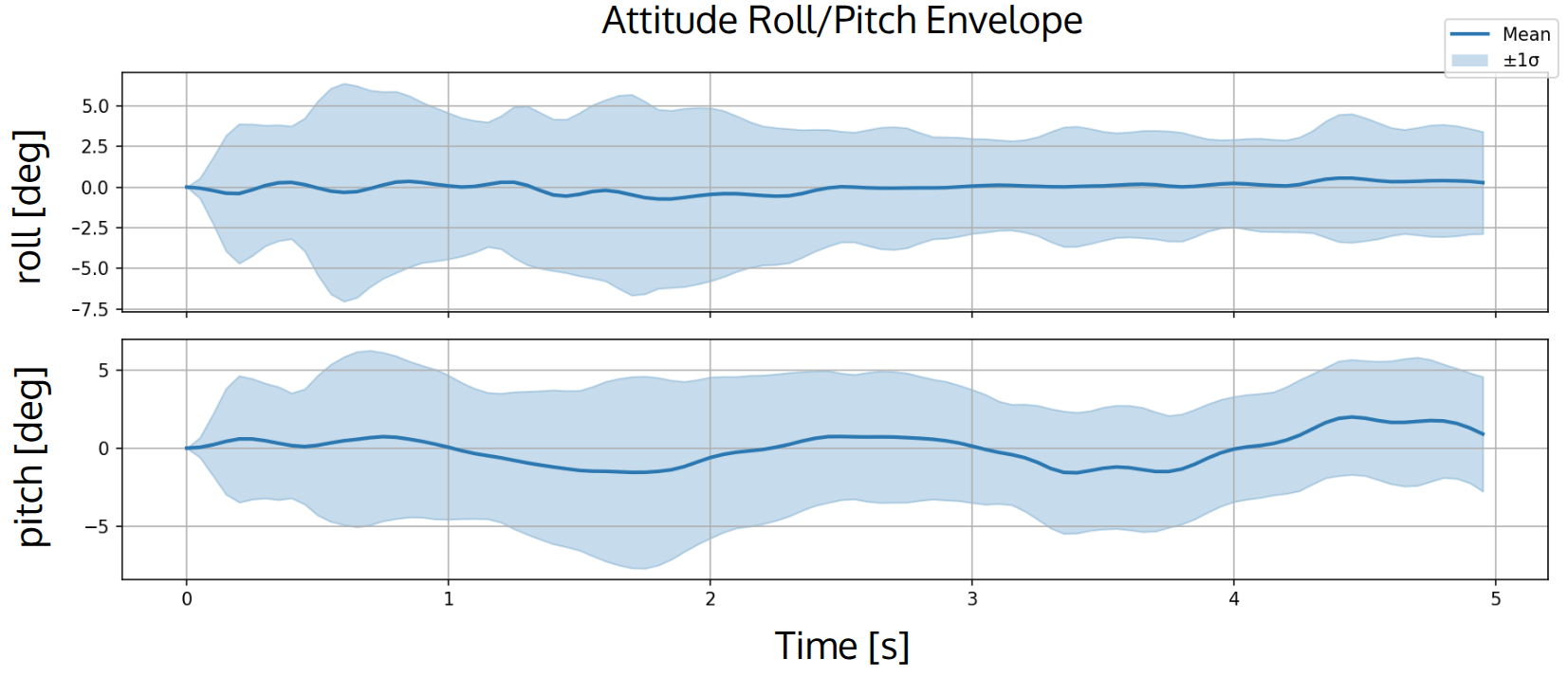}
    \end{minipage}\hfill
    \begin{minipage}[t]{0.48\textwidth}
        \centering
        Nominal MPC\\[2pt]
        \includegraphics[width=\linewidth]{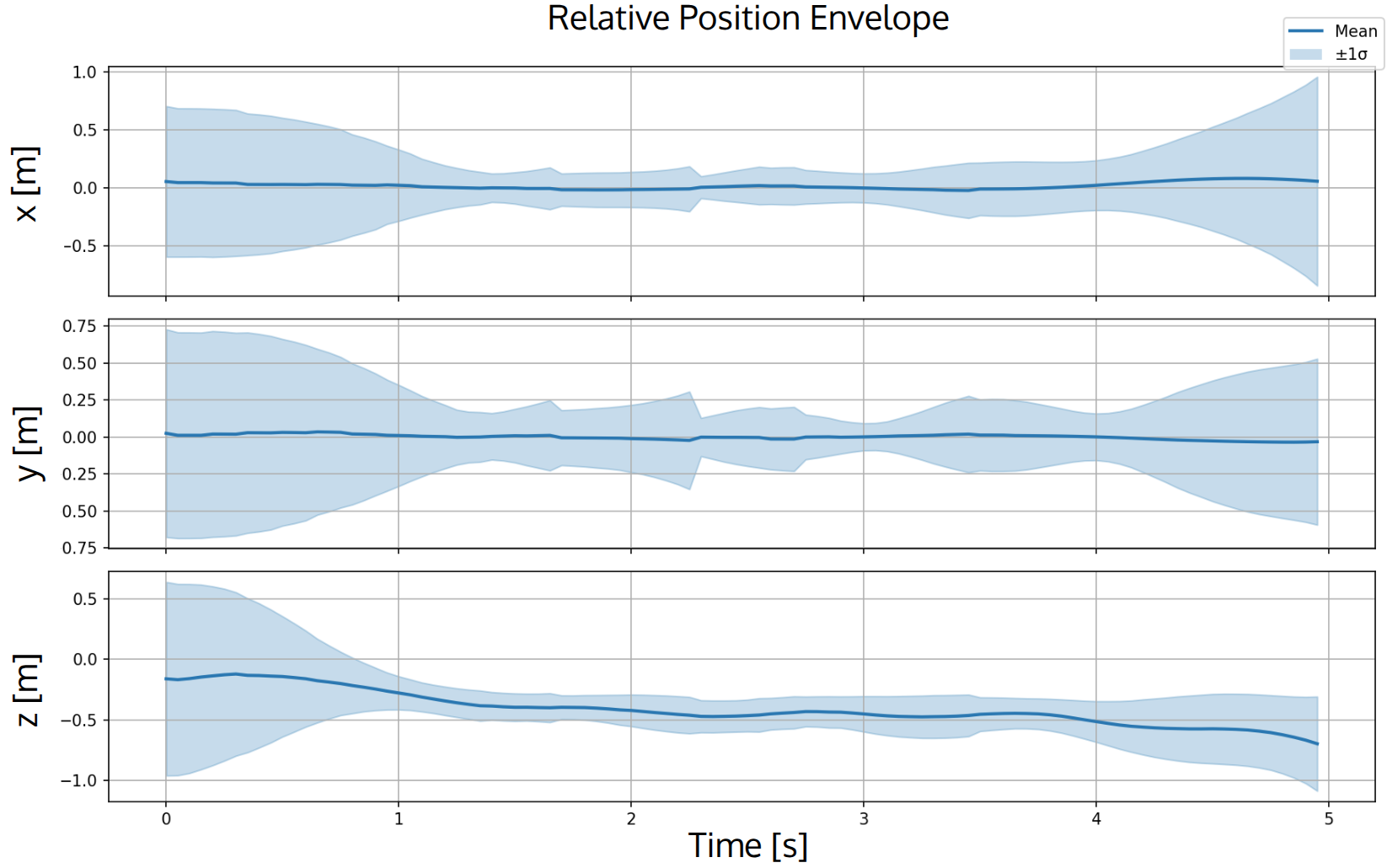}
        \includegraphics[width=\linewidth]{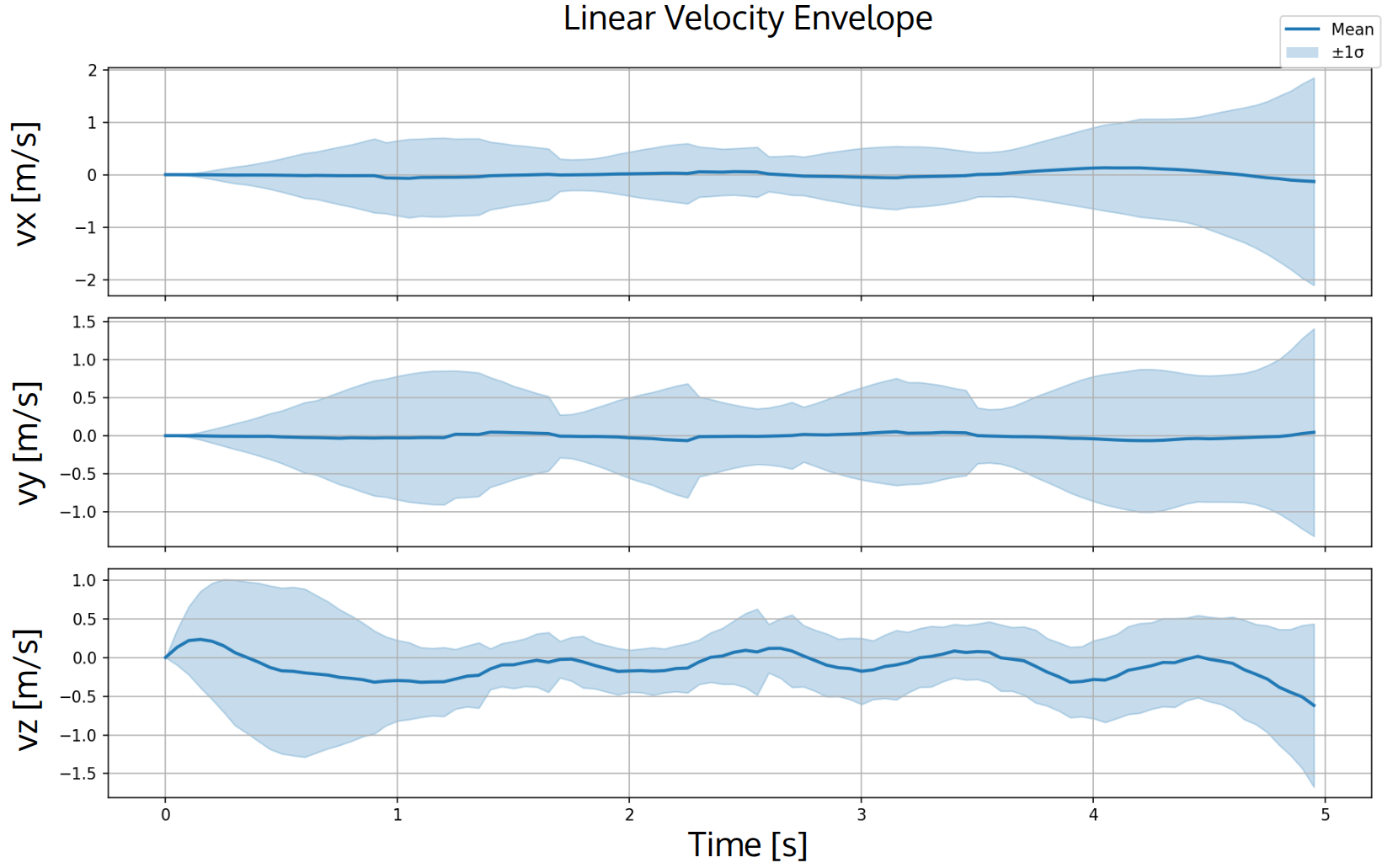}
        \includegraphics[width=\linewidth]{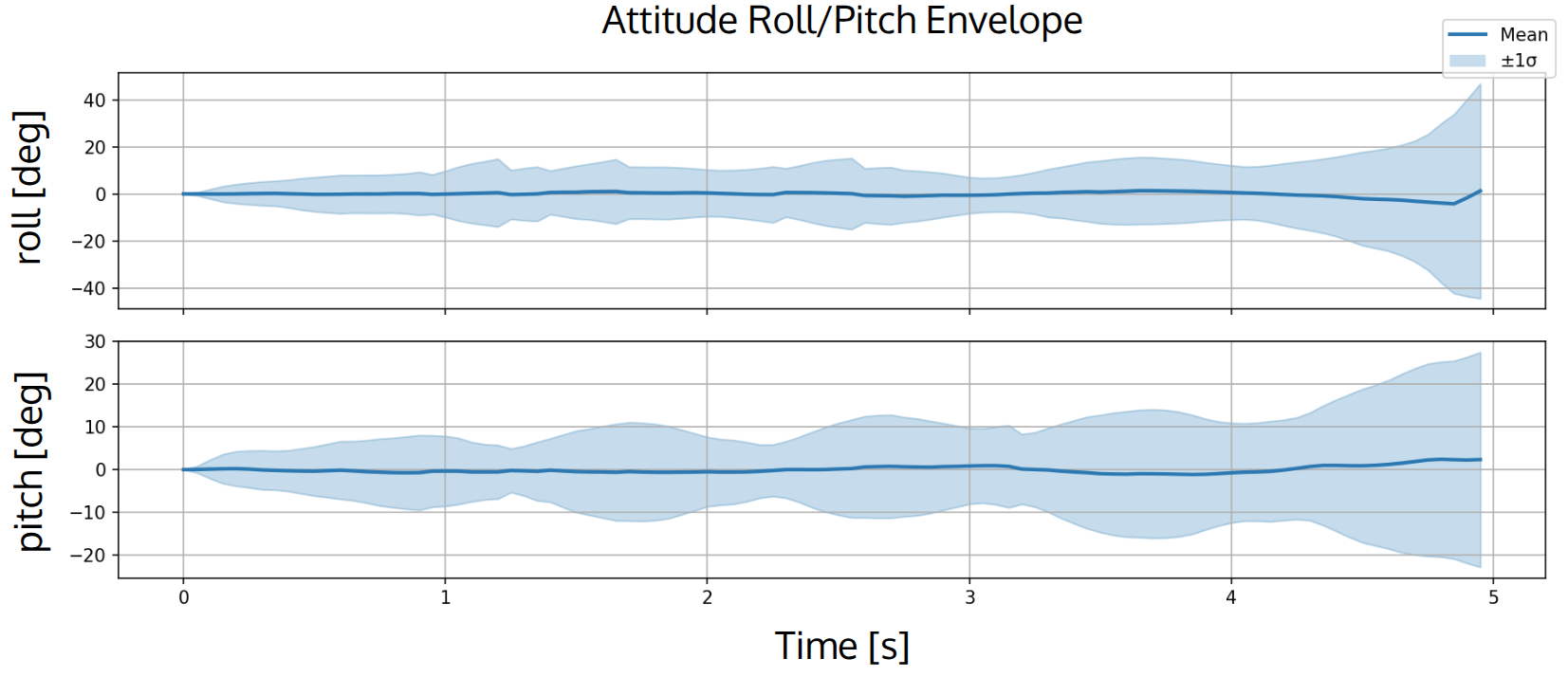}
    \end{minipage}
    \caption{Monte Carlo envelopes over $N=300$ randomized rollouts. Left: blended-critic MPC. Right: nominal MPC. The learned controller keeps the mean close to the hover target and prevents late-episode variance growth. The nominal controller develops large dispersion and a systematic loss of altitude, consistent with repeated crash trajectories.}
    \label{fig:quadrotor_mc_compare}
\end{figure*}

\paragraph{Monte Carlo analysis.}
The single-rollout picture is confirmed by the Monte Carlo envelopes in Figure~\ref{fig:quadrotor_mc_compare}, computed from $N=300$ randomized closed-loop runs. Under the learned controller, the mean trajectories converge toward hover and the state dispersion contracts after the initial transient. Under nominal MPC, the dispersion grows late in the episode as the quadrotor tips over and crashes during breaking. The nominal mean also develops a clear downward drift in $z$. This is the aggregate version of the same failure mode seen in Figure~\ref{fig:quadrotor_rollout_compare}.

\begin{figure*}[t]
    \centering
    \includegraphics[width=\linewidth]{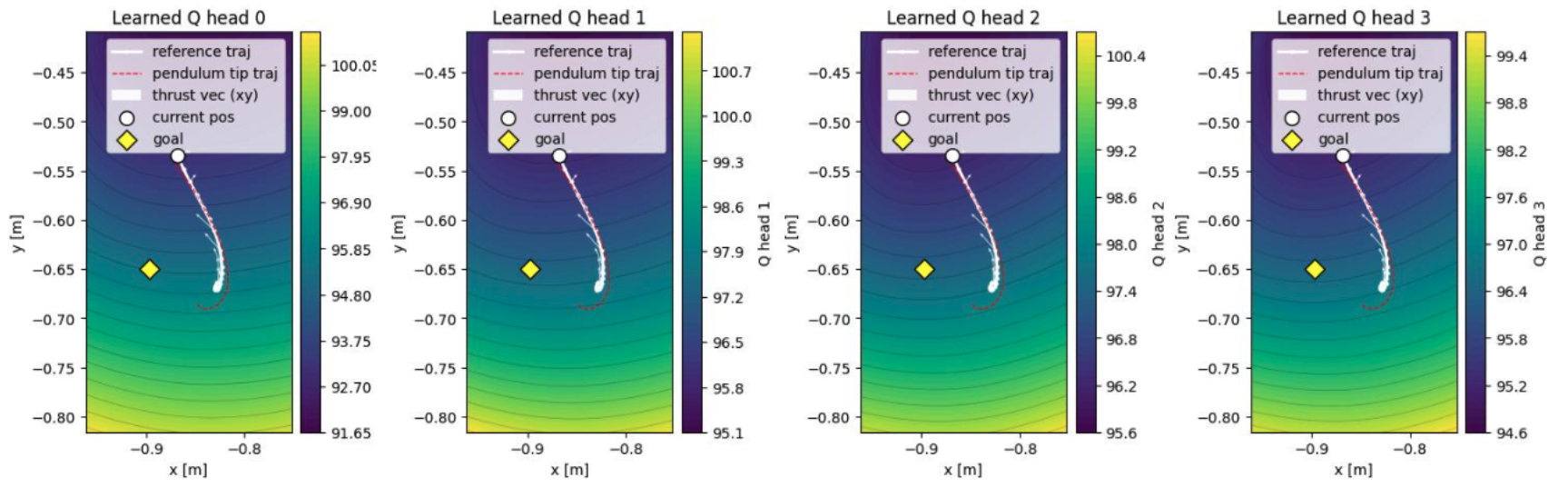}
    \caption{Local $x$-$y$ slices of the four learned value heads at one MPC optimization snapshot. All state and action entries except planar position are held fixed. The heads agree on the spatial value distribution, indicating low ensemble uncertainty for the critic correction at this rollout state.}
    \label{fig:quadrotor_q_heads}
\end{figure*}

\begin{figure*}[t]
    \centering
    \includegraphics[width=\linewidth]{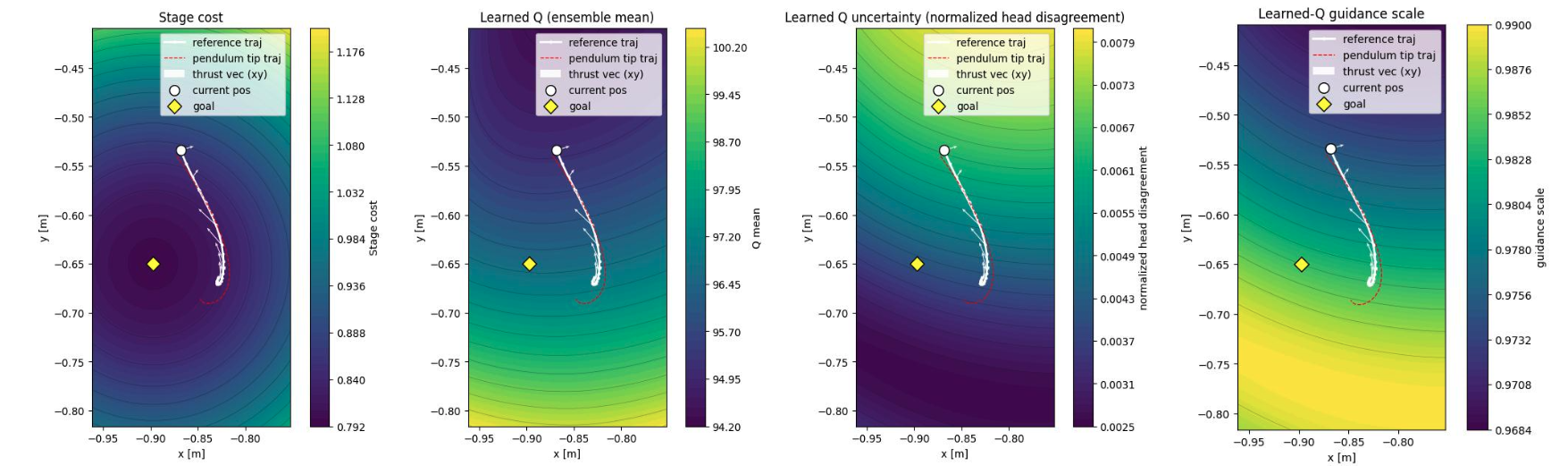}
    \caption{Local critic analysis for the same slice as Figure~\ref{fig:quadrotor_q_heads}. The learned value remains consistent with the stage cost by avoiding low-value regions far from the goal, but shifts the local optimum according to the fixed velocity, attitude, and payload state. The critic is smooth in the plotted region, and the value-head variance defines the guidance scale used to blend the critic with the nominal MPC objective.}
    \label{fig:quadrotor_q_scaling}
\end{figure*}

\paragraph{Local structure of the learned critic.}
Figures~\ref{fig:quadrotor_q_heads} and~\ref{fig:quadrotor_q_scaling} show local $x$-$y$ slices of the learned critic. Each slice is generated from one state-action sample taken during MPC optimization: only the quadrotor position coordinates $p_x$ and $p_y$ are swept, while the remaining state and action entries are fixed. The plots therefore describe the value density around this specific rollout condition; they are not global projections of the learned $Q$ function, which is fitted over the full $21$-dimensional state-action input. Figure~\ref{fig:quadrotor_q_heads} shows that the four value heads converge to almost identical spatial value distributions in this slice. The critic is therefore certain about the correction signal used at this rollout state.

Figure~\ref{fig:quadrotor_q_scaling} compares this learned correction to the nominal stage cost. The critic remains consistent with the stage cost in the important failure case: it does not create spurious low-value basins far away from the goal and remains locally smooth. At the same time, the value minimum is shifted relative to the instantaneous stage-cost minimum. For the fixed velocity, attitude, and payload state in this snapshot, positions closer to the goal have higher predicted continuation cost; the same non-position states would be preferable farther back along the approach direction. This is the correction to account for the increased momentum the nominal model cannot express. The critic surface is smooth over the plotted region, which is required for stable iLQR derivatives. The final panel shows the uncertainty-based scaling derived from value-head variance, which is used to weight the critic against the nominal MPC objective (cf. Section~\ref{par:value_ensembles}).

\section{Cube Stacking Experiment Details}
\label{app:stacking}

\subsection{Task Setup}
\label{app:stacking_experiment_setup}

\paragraph{Dynamics Model.}
The manipulation benchmark asks the robot to pick up a target cube and place it on top of a support cube. The planner uses the reduced state and control
\begin{equation*}
\begin{aligned}
x &= \begin{bmatrix} q_{\mathrm{arm}} \\ w_g \\ p_{WT}^W \\ g \end{bmatrix},
\qquad
u = \begin{bmatrix} \dot q_{\mathrm{arm}} \\ \dot w_g \end{bmatrix},
\qquad
\dot{x} = f(x,u) = \begin{bmatrix} \dot q_{\mathrm{arm}} \\ \dot w_g \\ g\, v_{WE}^W(q_{\mathrm{arm}}, \dot q_{\mathrm{arm}}) \\ 0 \end{bmatrix},
\end{aligned}
\end{equation*}
where $q_{\mathrm{arm}}$ are arm joints, $w_g$ is gripper width, $p_{WT}^W$ is target-cube position, and $g \in [0,1]$ is a learned continuous grasp gate. Thus the nominal model transports the cube kinematically only in proportion to the current grasp estimate: for $g \approx 0$ the target cube is effectively decoupled from the end effector, whereas for $g \approx 1$ it follows the commanded end-effector motion.

The end-effector translational velocity is computed from differentiable forward kinematics of the manipulator. Let $p_{WE}^W = \mathrm{FK}_E(q_{\mathrm{arm}})$ denote the end-effector position obtained from the robot's batched forward-kinematics model. The corresponding geometric Jacobian is
\begin{equation*}
J_E^W(q_{\mathrm{arm}}) = \frac{\partial p_{WE}^W}{\partial q_{\mathrm{arm}}},
\qquad
v_{WE}^W(q_{\mathrm{arm}}, \dot q_{\mathrm{arm}}) = J_E^W(q_{\mathrm{arm}}) \dot q_{\mathrm{arm}}.
\end{equation*}
In the implementation, the forward kinematics and geometric Jacobians are evaluated in batch across all environments and are fully differentiable. Contact dynamics are not modeled explicitly; instead, the missing interaction effects must be absorbed by the learned residual and the learned value function.

\paragraph{Cost Function.}
Let $p_{WA}^W$ denote the support-cube center and define the desired stacked target position $p_{WG}^W = p_{WA}^W + [0, 0, 1.2 s_z]^\top$, with $s_x$ and $s_z$ the cube width and height. We implement a multi-phase cost function, where the pregrasp-term and the transport term are weighted by the grasping gate
\begin{equation*}
\begin{aligned}
\ell(x,u) ={}& \beta_p \lVert p_{WT}^W - p_{WE}^W \rVert_2 + \beta_p \, \operatorname{clip}(1.8 s_x - p_{WT,z}^W, 0, 1.8 s_x) \\
&+ g\, \beta_p \lVert p_{WG}^W - p_{WE}^W \rVert_2 + (1-g) \, \operatorname{sg}\!\left[\beta_p \lVert p_{WG}^W - p_{WE}^W \rVert_2\right] \\
&- \beta_g \Bigl( \rho(p_{WT}^W) \, \dot w_g - \bigl(1-\rho(p_{WT}^W)\bigr) g \, \dot w_g \Bigr) + \beta_u \lVert \dot q_{\mathrm{arm}} \rVert_2^2 + \beta_g \dot w_g^2,
\end{aligned}
\end{equation*}
where $\beta_p$, $\beta_g$, and $\beta_u$ are positive cost weights, $\rho(p_{WT}^W) = \exp\!\bigl(-\lVert p_{WG}^W - p_{WT}^W \rVert_2 / (0.4 s_x)\bigr)$, and $\operatorname{sg}[\cdot]$ denotes the stop-gradient evaluation used in the implementation.

\paragraph{Preprocessed Inputs.}
The residual dynamics model receives the feature vector
\begin{equation*}
\phi_f = \begin{bmatrix} x^\top & u^\top & (f_{L}^W)^\top & (f_{R}^W)^\top & g & z_{\mathrm{img}}^\top \end{bmatrix}^\top,
\end{equation*}
that is, the full state, control, left and right finger contact forces, the scalar grasp score, and optional image features $z_{\mathrm{img}}$. The value model receives
\begin{equation*}
\phi_Q = \begin{bmatrix} (p_{WG}^W)^\top & x^\top & u^\top & (f_{L}^W)^\top & (f_{R}^W)^\top & z_{\mathrm{img}}^\top \end{bmatrix}^\top,
\end{equation*}
so it additionally gets the goal location but does not take the grasp score as a separate input in the current implementation.

\subsection{Experimental Insights}
\label{app:stacking_experiment_insights}

The ablations in Tables~\ref{tab:influence_of_horizon}, \ref{tab:influence_of_input}, and~\ref{tab:influence_of_dynamicsLearning} support the findings summarized in Section~\ref{sec:cube_stacking_experiment}. On this task, the main benefit comes from critic learning with informative observations, whereas residual dynamics learning yields only moderate improvement.

\paragraph{Effect of planning horizon.}
Table~\ref{tab:influence_of_horizon} shows that nominal MPC works best with a medium length optimization horizon. Very short horizons do not result in a sufficiently strong grasp, keeping the robot stuck in the approach phase, while very long horizons degrade due to unsteadiness in the iLQR backward pass when optimizing over the grasp. The learned critic improves success across the full horizon range and already reaches $0.685$ success rate at $H=5$, where nominal MPC barely succeeds. This supports the main-text claim that the critic gives iLQR consistently a descent direction aligned with the long-horizon stacking objective, leading to more directed behavior and  more robust grasping. The learned critic also degrades at long horizons; unlike the exact critic of Section~\ref{par:lambda_study}, it supplies imperfect gradients and curvature at every stage, while the weight of the nominal rollout decays as $(\gamma\lambda)^k$, and both effects can destabilize the backward pass over long horizons (cf. Section~\ref{sec:limitations}).

\begin{table}[t]
    \centering
    \caption{Influence of iLQR horizon length on cube-stacking performance at $\Delta t_{\mathrm{ctrl}}=0.05\,\mathrm{s}$. Nominal MPC requires an intermediate horizon, whereas the learned critic maintains strong performance even for short horizons.}
    \label{tab:influence_of_horizon}
    \begingroup
    \begin{tabular}{lcccc}
        \hline
        $H$ & Nominal SR & Nominal cost & Critic SR & Critic cost \\
        \hline
        2  & 0.00  & --  & 0.03 & 278 \\
        5  & 0.116 & 280 & 0.685 & 266 \\
        10 & 0.318 & 254 & 0.885 & 223 \\
        20 & 0.361 & 251 & 0.941 & 220 \\
        30 & 0.298 & 253 & 0.018 & 335 \\
        50 & 0.0  & 411 & 0.02 & 421 \\
        \hline
    \end{tabular}
    \endgroup
\end{table}

\begin{figure}[t]
    \centering
    \includegraphics[width=\linewidth]{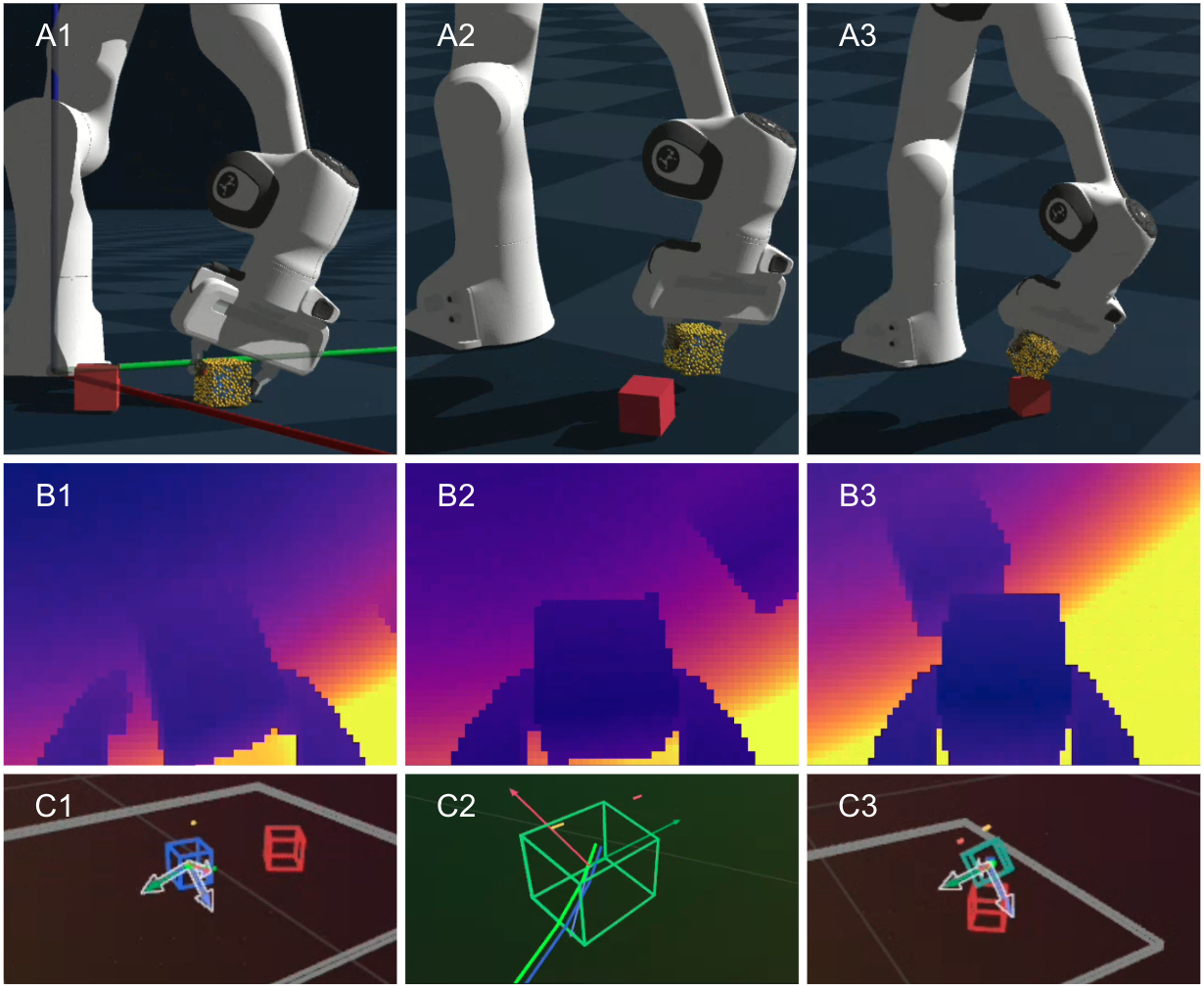}
    \caption{Cube-stacking phases. (1) Approach and Grasp; (2) Pickup and Transport; (3) Align and Drop; Depth (B) and haptic observations (contact-probes in gripper fingers visualized by red markers, figure A1..3) improve grasp estimation. The predicted cube trajectory (C2, blue) matches the predicted end-effector trajectory (C2, green) during the transport phase (2).}
    \label{fig:depth}
\end{figure}

\paragraph{Effect of visual and haptic inputs.}
Table~\ref{tab:influence_of_input} shows a consistent gain when depth and haptic observations are added to the critic input. Using only state, action, and goal features yields a success rate of $0.23$. Adding depth raises this to $0.543$, and adding haptic information further raises it to $0.941$. Figure~\ref{fig:depth} illustrates the transport phase together with the predicted cube and end-effector trajectories available to the controller. This matches the main-text result: visual and haptic observations stabilize the critic by reducing ambiguity about grasp quality and task phase, which is the dominant source of uncertainty in this benchmark.

\begin{table}[t]
    \centering
    \caption{Influence of critic input features on cube-stacking performance. Adding depth and haptic observations improves success rate monotonically, indicating that richer observations improve grasp-phase inference and critic stability.}
    \label{tab:influence_of_input}
    \begingroup
    \footnotesize
    \setlength{\tabcolsep}{4pt}
    \begin{tabular}{lcc}
        \hline
        Critic input & SR & Mean cost \\
        \hline
        $[x,u,p_{\mathrm{goal}}]$ & 0.230 & 238.0 \\
        $[x,u,p_{\mathrm{goal}},\mathrm{depth}]$ & 0.543 & 234.0 \\
        $[x,u,p_{\mathrm{goal}},\mathrm{depth},\mathrm{haptic}]$ & 0.941 & 257.0 \\
        \hline
    \end{tabular}
    \endgroup
\end{table}

\paragraph{Effect of residual dynamics learning.}
Table~\ref{tab:influence_of_dynamicsLearning} shows only minor benefit from learning residual dynamics on this task. The learned variants remain close to or below the baseline success rate and consistently increase the mean cost. This supports the hypothesis stated in the results section: modeling the exact dynamics of the cube is not the bottleneck in this task. Once the task phase is inferred robustly, the phased nominal model is already sufficient to plan approach, lift, and transport, whereas additional residual dynamics introduces optimization non-convexity that hurts iLQR. In practice, the added residual model occasionally induced erratic motions, such as brief gripper opening or slight end-effector jerks. During the transport phase, when the end effector had to maintain a firm grasp, such events often led to dropping the object, reducing success rate and increasing cost.

\begin{table}[t]
    \centering
    \caption{Influence of residual dynamics learning on cube-stacking performance. Residual variants do not improve significantly over the nominal baseline and often increase cost.}
    \label{tab:influence_of_dynamicsLearning}
    \begingroup
    \footnotesize
    \setlength{\tabcolsep}{3pt}
    \begin{tabular}{p{0.54\linewidth}ccc}
        \hline
        Experiment & SR & Mean cost \\
        \hline
        Baseline & 0.394 & 234.7 \\
        $\dot g \leftarrow [x,u,\mathrm{contact}]$ & 0.341 & 275.6 \\
        $\dot g \leftarrow [\mathrm{contact},\mathrm{depth}]$ & 0.401 & 278.2 \\
        $\dot g \leftarrow [x,u,\mathrm{contact},\mathrm{depth}]$ & 0.323  & 273.4 \\
        $\dot x \leftarrow [x,u,\mathrm{contact}]$ & 0.361  & 296.4 \\
        \hline
    \end{tabular}
    \endgroup
\end{table}

\section{Solver and Learning Settings}
\label{app:solver_and_learning_settings}

Tables~\ref{tab:shared_experiment_settings},~\ref{tab:task_specific_experiment_settings}, and~\ref{tab:ppo_settings} summarize the shared solver and learning configuration, the task-specific settings used in the quadrotor and cube-stacking experiments, and the PPO hyperparameters.

\begin{table}[h]
    \centering
    \caption{Shared solver, optimization, and data-collection settings used in both tasks.}
    \label{tab:shared_experiment_settings}
    \begingroup
    \footnotesize
    \setlength{\tabcolsep}{4pt}
    \begin{tabular}{p{0.48\linewidth}p{0.38\linewidth}}
        \hline
        Setting & Value \\
        \hline
        Planning time step $\Delta t_{\mathrm{plan}}$ & $0.05\,\mathrm{s}$ \\
        Horizon $H$ & $20$ \\
        Parallel environments & $100$ \\
        iLQR iterations & $8$ \\
        Maximum line-search trials & $9$ \\
        Maximum regularization updates & $10$ \\
        Executed control steps per replan & $3$ \\
        Residual-model initialization & Gaussian, standard deviation $10^{-3}$ \\
        Residual rollout loss & $10$ steps, weight $0.1$, midpoint integration \\
        Critic initialization & Gaussian, standard deviation $10^{0}$ \\
        Value-target construction & TD(5) \\
        Variance-based $\lambda$ attenuation & $\kappa = 4.0$ \\
        Blend schedule & $1.0 \rightarrow 0.9$ \\
        Update interval & every $20$ control steps \\
        Transitions per update & $2 \times 10^3$ \\
        Mini-batches per update & $10$ \\
        Critic replay buffer & $10^4$ \\
        Dynamics replay buffer & $10^7$ \\
        \hline
    \end{tabular}
    \endgroup
\end{table}

\begin{table}[h]
    \centering
    \caption{Task-specific timing and network settings.}
    \label{tab:task_specific_experiment_settings}
    \begingroup
    \footnotesize
    \setlength{\tabcolsep}{4pt}
    \begin{tabular}{p{0.34\linewidth}p{0.25\linewidth}p{0.25\linewidth}}
        \hline
        Setting & Quadrotor & Cube stacking \\
        \hline
        Episode length & $5\,\mathrm{s}$ & $8\,\mathrm{s}$ \\
        Control time step $\Delta t_{\mathrm{ctrl}}$ & $0.005\,\mathrm{s}$ & $0.01\,\mathrm{s}$ \\
        Residual model & ELU MLP, $3$ layers, width $128$ & ELU MLP, $3$ layers, width $256$ \\
        Critic trunk & ELU MLP, $3$ layers, width $128$ & ELU MLP, $3$ layers, width $256$ \\
        Value ensemble & $4$ heads, $1$ layer, width $128$, mean aggregation & $4$ heads, $1$ layer, width $256$, mean aggregation \\
        \hline
    \end{tabular}
    \endgroup
\end{table}

\begin{table}[h]
    \centering
    \caption{PPO hyperparameters.}
    \label{tab:ppo_settings}
    \begingroup
    \footnotesize
    \setlength{\tabcolsep}{4pt}
    \begin{tabular}{p{0.48\linewidth}p{0.38\linewidth}}        
        \hline
        Setting & Value \\
        \hline        
        Parallel Environments & $10^4$ \\
        Learning epochs & $5$ \\
        Mini-batches & $4$ \\
        Clipping parameter & $0.2$ \\        
        Discount factor $\gamma$ & $0.99$ \\
        GAE parameter $\lambda$ & $0.95$ \\
        Value-loss coefficient & $1.0$ \\
        Entropy coefficient & $0.01$ \\
        Learning rate & $10^{-3}$ \\        
        Desired KL & $0.01$ \\
        \hline
    \end{tabular}
    \endgroup
\end{table}

\section{Quaternion Projection}
\label{app:quat}
Often in aerospace and robotics, one is interested in optimizing over singularity-free rotation parametrizations. To make the solver accessible to a broad range of tasks, we require native support for state vectors that mix Euclidean states with quaternion components. During the forward rollout, quaternion states are renormalized after numerical integration, and during line search, the displacement $\mathrm{d}x(\hat x_i,x_i)$ uses Euclidean differences on vector blocks and local rotation errors on quaternion blocks. In the backward pass, quaternion derivatives are projected to local tangent coordinates because direct perturbations in the four-dimensional quaternion coordinates introduce radial components that violate $\|q\|=1$ and make the local quadratic model inconsistent after renormalization~\cite{Jackson2021}. We therefore map derivatives from the quaternion representation $q$ to a local angle-axis coordinate $\phi\in\mathbb{R}^3$ using the geometric Jacobian $G(q)$. For a quaternion-dependent function $h$, this gives $h_x^\mathrm{red}=E^\top h_x$ and $h_{xx}^\mathrm{red}\approx E^\top h_{xx}E$, where the Hessian projection omits second-order curvature terms in the spirit of the iLQR Gauss-Newton approximation. For dynamics, input and output states lie in different tangent spaces, so $f_{x,i}^\mathrm{red}=E_{i+1}^\top f_{x,i}E_i$ and $f_{u,i}^\mathrm{red}=E_{i+1}^\top f_{u,i}$~\cite{Jackson2021}. Appendix~\ref{app:quadrotor_experiment_setup} gives a concrete quadrotor example, and in the following we will detail the construction of $G(q)$.

Given a current unit quaternion $q$ and a local Rodrigues perturbation $\psi\in\mathbb{R}^3$, define
\begin{equation*}
\delta q(\psi)=\frac{1}{\sqrt{1+\|\psi\|^2}}
\begin{bmatrix}
1\\
\psi
\end{bmatrix},
\qquad
q^+(\psi)=q\otimes \delta q(\psi).
\end{equation*}
For a function $h:S^3\to\mathbb{R}^p$, the local rotational Taylor expansion is
\begin{equation*}
h(q^+(\psi))
=
h(q)+
\frac{\partial h(q\otimes \delta q(\psi))}{\partial \psi}\bigg|_{\psi=0}\psi
+ \mathcal{O}(\|\psi\|^2).
\end{equation*}
Using left quaternion multiplication $q\otimes \delta q=L(q)\delta q$, the projected Jacobian is
\begin{equation*}
\frac{\partial h}{\partial \psi}
=
\frac{\partial h}{\partial q}
L(q)
\frac{\partial \delta q}{\partial \psi}\bigg|_{\psi=0}.
\end{equation*}
For the Cayley parametrization,
\begin{equation}
    \label{eq:quat_projection_matrix}
\frac{\partial \delta q}{\partial \psi}\bigg|_{\psi=0}
=
\begin{bmatrix}
0_{1\times 3}\\
I_3
\end{bmatrix}
\quad\text{and therefore}\quad
G(q)=L(q)\begin{bmatrix}
0_{1\times 3}\\
I_3
\end{bmatrix},
\end{equation}
which gives the result in~\cite{Jackson2021}.

with $q=[w, v^\top]^\top$ and quaternion left-multiplication matrix
\begin{equation*}
L(q)=
\begin{bmatrix}
w & -v^\top\\
v & w I_3 + [v]_\times
\end{bmatrix}.
\end{equation*}
For the Hessian, we reuse the first-order projection matrix from Eq.~\eqref{eq:quat_projection_matrix}, yielding the projected Hessian (without 2nd-order curvature)
\begin{equation*}
h_{\phi\phi} = G(q)^\top h_{qq}(q)G(q).
\end{equation*}

\clearpage
\bibliography{references}

\end{document}